\documentclass{article}
\usepackage[utf8]{inputenc}
\usepackage{algorithm}
\usepackage{algpseudocode}
\usepackage{authblk}
\usepackage[top=1in, bottom=1in, left=1in, right=1in]{geometry}
\usepackage{tikz}
\usetikzlibrary{shapes,arrows,matrix,patterns,positioning}
\usepackage{color}
\usepackage{graphicx}
\usepackage{amssymb}
\usepackage{epstopdf}

\usepackage{float}
\usepackage{amsmath,amsthm,bm,color,epsfig,enumerate,caption}
\usepackage{hyperref}
\usepackage{booktabs}
\usepackage{multirow}
\usepackage{array}
\usepackage{xcolor}

\title{A Distributed Block Chebyshev-Davidson Algorithm for Parallel Spectral Clustering}
\author[1]{Qiyuan Pang}
\author[2]{Haizhao Yang \thanks{corresponding author, email: hzyang@umd.edu}}
\affil[1]{Department of Mathematics, Purdue University, West Lafayette}
\affil[2]{Department of Mathematics and Department of Computer Science, University of Maryland, College Park}
\date{}

\begin{document}

\maketitle

\begin{abstract}
We develop a distributed Block Chebyshev-Davidson algorithm to solve large-scale leading eigenvalue problems for spectral analysis in spectral clustering. First, the efficiency of the Chebyshev-Davidson algorithm relies on the prior knowledge of the eigenvalue spectrum, which could be expensive to estimate. This issue can be lessened by the analytic spectrum estimation of the Laplacian or normalized Laplacian matrices in spectral clustering, making the proposed algorithm very efficient for spectral clustering. Second, to make the proposed algorithm capable of analyzing big data, a distributed and parallel version has been developed with attractive scalability. The speedup by parallel computing is approximately equivalent to $\sqrt{p}$, where $p$ denotes the number of processes. {Numerical results will be provided to demonstrate its efficiency in spectral clustering and scalability advantage over existing eigensolvers used for spectral clustering in parallel computing environments.} 
\end{abstract}


\section{Introduction}
Spectral clustering has a long history \cite{shi2000normalized, cheeger2015lower, donath1972algorithms, fiedler1973algebraic, guattery1994performance, spielman1996spectral, ng2001spectral} and it was popularized as a machine learning model by Shi \& Malik \cite{shi2000normalized} and Ng, Jordan, \& Weiss \cite{ng2001spectral}.
Spectral clustering makes use of the spectrum of the similarity matrix of the data to perform dimensionality reduction before clustering in fewer dimensions. The basic algorithm is summarized as follows:

\begin{itemize}
    \item[] 1. Calculate the symmetric normalized Laplacian $A$ of an undirected graph with $N$ graph nodes.
    \item[] 2. Compute the first $k$ eigenvectors corresponding to the smallest $k$ eigenvalues of $A$.
    \item[] 3. {Normalize each row of the eigenvectors and use the resulting matrix as the feature matrix where the $l$-th row defines the features of graph node $l$.}
    \item[] 4. Cluster the graph based on the features using clustering methods like K-means.
\end{itemize}

{Throughout the paper, we consider the symmetric normalized Laplacian $A \in \mathbb{R}^{N\times N}$ of a undirected graph of $N$ nodes:
\begin{equation}
    A = I - D^{-1/2}S D^{-1/2},
\end{equation}
where $I$ is the identity matrix, $S$ is the similarity matrix of the graph with $S_{ij} = 1$ if nodes $i$ and $j$ are connected otherwise $0$, and $D_{ii} = \sum_{j}S_{ij}$ is the digonal degree matrix. We always assume $A$ is sparse because most graphs in real applications are sparse. We focus on solving the large-scale leading eigenvalue problem, which is the second step of spectral clustering. Given the symmetric normalized Laplacian matrix $A$, the leading eigenvalue problem is to find
\begin{equation}
\label{eq:eigproblem}
    A V = V \Lambda_{k},
\end{equation}
where $\Lambda_{k} \in \mathbb{R}^{k\times k}$ is a diagonal matrix with $0 \leq \lambda_{1} \leq \lambda_{2} \leq ... \leq \lambda_{k} \leq 2$ as the $k$ smallest eigenvalues of $A$ along the diagonal, and  $V \in \mathbb{R}^{N\times k}$ is a tall-skinny ($k << N$) matrix consisting of the corresponding eigenvectors.}

The Chebyshev-Davidson method is a good candidate to solve the eigenproblem with fast convergence for spectral clustering, since the spectrum bounds of the Laplacian and normalized Laplacian matrices in spectral clustering are known explicitly \cite{cheeger2015lower}. The Chebyshev-Davidson method was first introduced \cite{zhou2007chebyshev} as an eigensolver for quantum chemistry systems \cite{szabo2012modern, lu2017preconditioning, li2017interior, zhou2014chebyshev} and symmetric eigenvalue problems \cite{saad2011numerical, schofield2012using, zhouaccelerating, miao2017filtered}, which applies a Chebyshev polynomial filter to accelerate the convergence of the Davidson method \cite{crouzeix1994davidson}. 
The Davidson method can augment the searching subspace for eigenvectors by a potentially better new vector than the one based on a strict Krylov subspace structure, resulting in faster convergence. The augmentation vector added to the subspace at each step requires solving a correction equation that is not affordable for big data, even though the Jacobi-Davidson method \cite{sleijpen2000jacobi} has been designed to favor the efficient use of modern iterative techniques for the correction-equation, based on preconditioning and Krylov subspace acceleration. Compared to all other types of Davidson-type methods, there is no need to form or solve any correction equations in the Chebyshev-Davidson method. Instead, interval-wise filtering based on Chebyshev polynomials is utilized, making it very suitable for large-scale sparse matrices with known bounds of the spectrum. The Chebyshev filter can enhance the eigensubspace of interest and dampen the eigensubspace undesired, making the Chebyshev-Davidson method efficiently applicable to general matrices, including the sparse symmetric matrices in spectral analysis. The Chebyshev-Davidson method could be extended to the Block Chebyshev-Davidson method \cite{zhou2010block, teng2016block} with an inner-outer restart technique to reduce total CPU time and a progressive polynomial filter to take advantage of suitable initial vectors when they are available. Good initials are available and essential in many data science scenarios. For example, when partitioning a streaming graph changing over time using spectral clustering, eigenpairs computed for the previous graph are good initials for evaluating the eigenpairs of the current graph. {The Block Chebyshev-Davidson method and some variants are widely used in problems including linear response eigenvalue problem \cite{zhou2016accelerating}, partial eigenvalue decomposition \cite{ji2011block, miao2023flexible, miao2020chebyshev}, generalized eigenvalue problems \cite{wang2022new}, supramolecular systems \cite{koehl2018large}, correlated eigenvalue problems \cite{di2013block} and self-consistent-filed calculations \cite{zhou2006parallel}. In this paper, we propose to use the Block Chebyshev-Davidson method for spectral clustering and develop a scalable version of the algorithm for parallel spectral clustering. The framework of spectral clustering via the Block Chebyshev-Davidson method is summarized as Algorithm \ref{algo:spclustering}.} 

Problems in data science areas like spectral clustering are usually large-scale, and parallel algorithms are a common practice to reduce CPU time. The multi-threading implementation for a single shared-memory node and the multi-processing/distributed implementation for distributed-memory nodes are standard parallelization techniques. Multi-threading algorithms are easy to implement, but their ability to handle large-scale problems and accelerate solutions is still restricted by the limited memory and number of threads in one single node. Multi-process algorithms are difficult to implement and scale due to the communication cost among processes; however, they could deal with large-scale matrices that cannot be stored or processed in a single node. Besides, with careful design, distributed algorithms could accelerate solutions much further. 

Since matrices in problem (\ref{eq:eigproblem}) are usually too large for one single node, even though they are sparse in spectral clustering in practice, we propose a novel multi-processing Block Chebyshev-Davidson method to solve the leading eigenvalue problem. There are three essential components in the Block Chebyshev-Davidson method to be parallelized in distributed memory: sparse times tall-skinny matrix multiplication (SpMM), Chebyshev polynomial filter to tall-skinny matrices, and tall-skinny matrix orthonormalization. A multi-processing implementation of the Block Chebyshev-Davidson method could be found in PARSEC package \cite{zhou2006parallel} and it is used to solve the Kohn-Sham eigenvalue problem \cite{kohn1965self, yu2018elsi, lu2017cubic}. However, it is not scalable to large concurrencies due to its 1D parallel SpMM algorithm in the Chebyshev polynomial filters and the parallel DGKS \cite{daniel1976reorthogonalization} for orthonormalization. Unlike the well-studied parallel algorithms for dense matrix-dense matrix multiplication (GEMM) \cite{cannon1969cellular, van1997summa, agarwal1995three, solomonik2011communication} and sparse matrix-sparse matrix multiplication (SpGEMM) \cite{azad2016exploiting, bulucc2012parallel, schatz2016parallel}, the research of parallel SpMM algorithms has not flourished until recent years due to the needs in data science areas. The flops in parallel GEMM scale with $N^3$ for multiplying two $N \times N$ matrices, whereas the communication costs scale with $N^2$. This benefit of parallel GEMM is called the surface-to-volume ratio. Parallel SpGEMM and SpMM do not benefit from this favorable computation-to-communication scaling. In fact, it is easier to scale SpGEMM than to scale SpMM to large concurrencies \cite{selvitopi2021distributed}. For example, the 1D SpMM algorithm scales poorly to dozens of concurrencies. Oguz Selvitopi et al. \cite{selvitopi2021distributed} compare A-Stationary 1.5D \cite{kannan2017mpi}, A-Stationary 2D, and C-Stationary 2D algorithms \cite{schatz2016parallel} and conclude that, unlike for GEMM and SpGEMM, 2D algorithms are not strongly scalable to large process counts for SpMM; instead, 1.5D achieves much better scaling. We adopt the A-Stationary 1.5D algorithm for SpMM and Chebyshev polynomial filter. In the A-Stationary 1.5D algorithm, the sparse matrix is partitioned in 2D while the tall-skinny matrix is partitioned in 1D. The resulting tall-skinny matrix is partitioned in 1D but in a different schema. Hence, the 1.5D algorithm generally does not apply to a Chebyshev polynomial filter which conducts multiple SpMMs and matrix-matrix additions sequentially. When the sparse matrix $A$ is symmetric and square, we could solve the issue by transposing the 2D process grid and re-distributing the tall-skinny matrices. 

Parallel orthonormalization is another issue because a parallel QR seldom scales due to communication costs. James Demmel et al. \cite{demmel2012communication} compare multiple parallel QR algorithms and present a parallel Tall Skinny QR (TSQR) algorithm, which attains known communication lower bounds and communicates as little as possible only up to polylogarithmic factors. The parallel TSQR algorithm leads to significant speedups in practice over some of the existing algorithms, including DGEQRF in LAPACK \cite{lapack99} and PFDGEQRF in ScaLAPACK \cite{slug}.
We adopt the parallel TSQR for orthonormalization, which is also more efficient than the parallel DGKS.
Though orthonormalization hardly scales to large concurrencies, orthonormalization only takes a small amount of computation cost compared to Chebyshev polynomial filters, especially when the ratio of the degree of the filter to the number of desired eigenvectors is large. Note that a higher ratio results in faster convergence as well. Therefore, the distributed Block Chebyshev-Davidson method is practically scalable and efficient for data science applications like spectral clustering. Table \ref{tb:complexity} summarizes the flops and communication costs of our distributed algorithm and the components.

Free software implementing spectral clustering is available in large open-source projects like Scikit-learn using LOBPCG \cite{knyazev2001toward} or ARPACK \cite{lehoucq1998arpack}, and MLlib for pseudo-eigenvector clustering using the power iteration method \cite{lin2010power}. Parallel versions of the eigensolvers mentioned are used in different parallel spectral clustering methods \cite{naumov2016parallel, chen2010parallel, yan2013p, huo2020designing}. Parallel LOBPCG is used in \cite{naumov2016parallel} for parallel spectral graph partitioning, and the speedup increases slowly as the number of processes becomes large. Numerical results in \cite{chen2010parallel, huo2020designing} using up to 256 processes show that the parallel ARPACK used in parallel spectral clustering is accelerated at a rate proportional to the square root $\sqrt{p}$ of the number of processes $p$. The parallel power iteration method in parallel Power Iteration Clustering (p-PIC) \cite{yan2013p} achieves linear speedups in one single node when the number of processes is smaller than 70, but then the speedups drop when the number of processes is larger. No numerical results are provided in \cite{yan2013p} to demonstrate the scalability when the number of processes is larger than 90. The speedup of a parallel eigensolver is usually high with dozens of processes but becomes less significant when the number of processes keeps growing. In most of the works mentioned above, the scalabilities of the parallel eigensolvers are tested with less than 256 processes. In Figure \ref{fig:scalability-arpack-lobpcg}, we test the scalability of parallel ARPACK and LOBPCG up to 1000 processes and show the lack of scalability when the number of processes is large. In our work, besides theoretical proof, we will test the scalability of our algorithm with up to 1000 processes to demonstrate that the speedup is approximately the square root $\sqrt{p}$ when the number of processes $p$ is large. See Figure \ref{fig:dbchdav-scaling}.

The contributions of this paper are summarized as follows:
\begin{itemize}
    \item We propose to use the Block Chebyshev-Davidson method as an efficient eigensolver for spectral clustering due to the known analytic spectrum bounds.
    \item {We develop a scalable distributed Block Chebyshev-Davidson method which achieves approximately $\sqrt{p}$ speedup when the number of processes $p$ is large. Our method is more scalable in parallel computing environments than the eigensolvers ARPACK and LOBPCG used in spectral clustering.}
\end{itemize}

\begin{algorithm}
\caption{Spectral Clustering via the Block Chebyshev-Davidson method}
\label{algo:spclustering}
\begin{algorithmic}[1]
\State \textbf{Input:} an undirected graph with $N$ graph nodes.

\State Calculate the symmetric normalized Laplacian $A$ of the undirected graph.
\State Use the Block Chebyshev-Davidson method to compute the first $k$ eigenvectors corresponding to the smallest $k$ eigenvalues of $A$.
\State Normalize each row of the eigenvectors and use the resulting matrix as the feature matrix where the $l$-th row defines the features of graph node $l$.
\State Cluster the graph based on the features using clustering methods like K-means.
\State \textbf{Output:} the cluster assignments of the graph nodes.
\end{algorithmic}

\end{algorithm}

\begin{table}[!htbp]
\caption{\textbf{Summary of the complexity per iteration of components in our distributed Block Chebyshev-Davidson algorithm for solving the problem (\ref{eq:eigproblem}) with $p$ processes. $k_{b}$ and $act_{max}$ ($k_{b} < act_{max}$) are constants related to $k $. $nnz(A)$ is the number of nonzero entries in $A$. See Section 3 for a detailed analysis.}}
\centering
\scalebox{0.85}{
\begin{tabular}{|c|c|c|c|}
\hline
\textbf{Components} & \textbf{$\#$Flops} & \textbf{$\#$Messages} & \textbf{$\#$Words} \\
\hline
Filter (deg $m$) & $O(\dfrac{nnz(A)m k_{b}}{p})$ & $O(m \log p)$ & $O(\dfrac{2mNk_{b}}{\sqrt{p}})$ \\
\hline
SpMM & $O(\dfrac{nnz(A) k_{b}}{p})$ & $O(\log p)$ & $O(\dfrac{2Nk_{b}}{\sqrt{p}})$ \\
\hline
Orthonormalization & $O(\dfrac{3Nact_{max}^2}{p} + 3act_{max}^3 \log p)$ & $O(\log p)$ & $O(act_{max}^2 \log p)$ \\
\hline
Update Rayleigh-quotient & $O(\dfrac{Nk_{b}act_{max}}{p})$ & $O(\log p)$ & $O(act_{max}k_{b}\log p)$ \\
\hline
Evaluate residual & $O(\dfrac{nnz(A)k_{b}+Nk_{b}^2}{p})$ & $O(\log p)$ & $O(\dfrac{2Nk_{b}}{\sqrt{p}})$ \\
\hline
Totals & $O(\dfrac{nnz(A)mk_{b}+N act_{max}^2}{p}+3act_{max}^3\log p)$ & $O(m\log p)$ & $O(\dfrac{2mNk_{b}}{\sqrt{p}}+act_{max}^2 \log p)$\\
\hline
\end{tabular}
}

\label{tb:complexity}
\end{table}

The rest of this paper is organized as follows. In Section 2, we will briefly introduce the Block Chebyshev-Davidson method. In Section 3, we will describe the distributed Block Chebyshev-Davidson. Numerical results are provided in Section 4 to demonstrate the effectiveness of our algorithm. Section 5 concludes the paper.

\section{Block Chebyshev-Davidson Method}
Chebyshev-Davidson \cite{zhou2007chebyshev} and Block Chebyshev-Davidson \cite{zhou2010block} algorithms employ Chebyshev polynomial filters to accelerate the convergence of the Davidson method. The latter applies an inner-outer restart technique inside an iteration that reduces the computational costs of using a large dimension subspace and a progressive filtering technique to take advantage of suitable initial vectors if available. The sequential Block Chebyshev-Davidson algorithm is summarized in Algorithm \ref{algo:bchdav-seq}, which is Algorithm 3.1 in \cite{zhou2010block}. {There are successful applications of the Block Chebyshev-Davidson method and its variants in problems including linear response eigenvalue problem \cite{zhou2016accelerating}, partial eigenvalue decomposition \cite{ji2011block, miao2023flexible, miao2020chebyshev}, generalized eigenvalue problems \cite{wang2022new}, supramolecular systems \cite{koehl2018large}, correlated eigenvalue problems \cite{di2013block} and self-consistent-filed calculations \cite{zhou2006parallel}. In this paper, we apply it to spectral clustering and develop a scalable parallel version of it in the next section. The spectral clustering via the Block Chebyshev-Davidson method is summarized as Algorithm \ref{algo:spclustering}.}

To understand the algorithm, we first introduce some input variables. $k_{want}$ denotes the number of eigenpairs one wants to evaluate. $k_{b}$ is the number of vectors added to the projection basis per iteration in Step 5 of Algorithm \ref{algo:bchdav-seq}. $k_{sub}$ is the dimension of the current subspace $V$, $k_c$ is the number of converged eigenvectors and $k_{act}$ is the dimension of the active subspace in $V$ deflated by the converged eigenvectors $V(:,1:k_{c})$. It always holds that $k_{c} + k_{act} = k_{sub}$. $act_{max}$ and $dim_{max}$ denote the maximum dimensions of the active subspace, and the subspace spanned by $V$, respectively. We will briefly introduce the essential components of the algorithm and refer the readers to the original paper \cite{zhou2010block} for more details.

\begin{algorithm}
\caption{Block Chebyshev-Davidson method with inner-outer restart}
\label{algo:bchdav-seq}
\begin{algorithmic}[1]
\State Compute or assign $upperb$, $lowb$, and $low\_nwb$.
\State Set $V_{tmp} = V_{init}(:,1:k_{b})$, (construct $k_{b}-k_{init}$ random vectors if $k_{init} < k_{b}$); set $k_{i} = k_{b}$($k_{i}$ counts the number of used vectors in $V_{init}$).
\State Set $k_{c} = 0, k_{sub} = 0$ ($k_{sub}$ counts the dimension of the current subspace); set $k_{act} = 0$ ($k_{act}$ counts the dimension of the active subspace.)
\While{ $itmax$ is not exceeded}
\State $V(:,k_{sub}+1:k_{sub}+k_{b}) = ChebyshevFilter(V_{tmp}, m, lowb, upperb, low\_nwb)$.
\State Orthonormalize $V(:,k_{sub}+1:k_{sub}+k_{b})$ against $V(:,1:k_{sub})$.
\State Compute $W(:,k_{act}+1:k_{act}+k_{b}) = AV(:,k_{sub}+1:k_{sub}+k_{b})$; set $k_{act}=k_{act}+k_{b}$; set $k_{sub} = k_{sub} + k_{b}$.
\State Compute the last $k_{b}$ columns of the Rayleigh-quotient matrix $H$: $H(1:k_{act},k_{act}-k_{b}+1:k_{act}) = V(:,k_{c}+1:k_{sub})^T W(:,k_{act}-k_{b}+1:k_{act})$, then symmetrize $H$.
\State Compute eigen-decomposition of $H(1:k_{act}, 1:k_{act})$ as $HY = YD$, where $diag(D)$ is in non-increasing order. Set $k_{old} = k_{act}$.
\State If $k_{act} + k_{b} > act_{max}$, then do inner restart as: $k_{act} = k_{ri}, k_{sub} = k_{act} + k_{c}$.
\State Do subspace rotation (final step of Rayleigh-Ritz refinement) as: $V(:,k_{c}+1:k_{c}+k_{act}) = V(:,k_{c}+1:k_{old})Y(1:k_{old}, 1:k_{act}), W(:,1:k_{act}) = W(:,1:k_{old})Y(1:k_{old}, 1:k_{act})$.
\State Compute residual $r = AV(:,k_{c}+1: k_{c}+k_{b}) - V(:,k_{c}+1: k_{c}+k_{b})D(k_{c}+1: k_{c}+k_{b},k_{c}+1:  k_{c}+k_{b})$ and test for convergence of the $k_{b}$ vectors in $V(:,k_{c}+1, k_{c}+k_{b})$, denote the number of newly converged Ritz pairs at this step as $e_{c}$. If $e_{c} > 0$, then update $k_{c} = k_{c} + e_{c}$, save converged Ritz values in $eval$(sort $eval(:)$ in non-increasing order), and deflate/lock converged Ritz vectors (only need to sort $V(:,1:k_{c})$ according to $eval(1:k_{c})$).
\State If $k_{c} \geq k_{want}$, then return $eval(1:k_{c})$ and $V(:,1:k_{c})$, exit.
\State If $e_{c} > 0$, set $W(:,1:k_{act}-e_{c}) = W(:,e_{c}+1:k_{act}), k_{act} = k_{act} - e_{c}$.
\State Update $H$ as the diagonal matrix containing non-converged Ritz values $H = D(e_{c}+1:e_{c}+k_{act}, e_{c}+1:e_{c}+k_{act})$.
\State If $k_{sub} + k_{b} > dim_{max}$, do outer restart as: $k_{sub} = k_{c} + k_{ro}, k_{act} = k_{ro}$.
\State Get new vectors for the next filtering: Set $V_{tmp} = [V_{init}(k_{i}+1:k_{i}+e_{c}), V(:,k_{c}+1:k_{c}+k_{b}-e_{c})]; k_{i} = k_{i} + e_{c}$.
\State Set $low\_nwb$ as the median of non-converged Ritz value in $D$.
\EndWhile
\end{algorithmic}

\end{algorithm}

\begin{algorithm}
\caption{$[W] = ChebyshevFilter(V, m, a, b, a_{0})$}
\label{algo:cheb-filter}
\begin{algorithmic}[1]
\State \textbf{variables}: $A$ the sparse matrix; $V$ the input matrix; $m$ the degree of a Chebyshev polynomial; $a$ the lower bound of unwanted eigenvalues of $A$; $b$ the upper bound of all eigenvalues; $a_{0}$ the lower bound of all eigenvalues; 

\State $c = (a + b) / 2$; $e = (b - a)/2$;
\State $\sigma = e/(a_{0}-c)$;
\State $t = 2/\sigma$;
\State $U = (AV - cV)\sigma/e $;
\For{$i = 2:m$}
\State $\sigma_{1} = 1/(\tau - \sigma)$;
\State $W = 2\sigma_{1}(A U - c U)/e - \sigma\sigma_{1}v$;
\State $V = U$;
\State $U = W$;
\State $\sigma = \sigma_{1}$;
\EndFor
\end{algorithmic}

\end{algorithm}

\textbf{Chebyshev polynomial filter (Step 5 of Algorithm \ref{algo:bchdav-seq}).} Given a full eigendecomposition $A = V \Lambda V^{T}$ of the symmetric matrix in the leading eigenvalue problem (\ref{eq:eigproblem}), for any polynomial $\phi(x): \mathbb{R} \hookrightarrow \mathbb{R}$, it holds 
\begin{equation}
\label{eq:polyA}
    \phi(A) = V \phi(\Lambda) V^{T}.
\end{equation}
The Chebyshev polynomial of degree $m$ is defined as 
\begin{equation}
\label{eq:chebypoly}
    C_{m}(x) = \left\{
        \begin{array}{ll}
            \cos(m \cos^{-1}(x)), & \quad -1 \leq x \leq 1, \\
            \cosh(m \cosh^{-1}(x)), & \quad |x| > 1,
        \end{array}
    \right.
\end{equation}
which rapidly grows outside the interval $[-1, 1]$. By (\ref{eq:polyA}), if $\phi(x) = C_{m}(s x+t)$ with appropriate $s$ and $t$, then the smallest $k$ eigenvalues of $A$ become the largest $k$ eigenvalues of $\phi(A)$ which are significantly much larger than other eigenvalues, making they well separated from others, which means that it is relatively much easier to identify the leading $k$ eigenpairs of $\phi(A)$ using the Davidson method without correction-equation. Note that the eigenvectors of $A$ and $\phi(A)$ remain unchanged by (\ref{eq:polyA}). Therefore, it is sufficient to find the $k$ largest eigenpairs of $\phi(A)$. Another advantage of Chebyshev polynomials is that they admit a three-term recurrence relation such that $\phi(A)$ can be efficiently applied to an arbitrary vector $v$ as long as the fast matvec of $A$ is available. For example, $\phi(A) = C_{m}(s A + t I_{N})$ can be applied to an arbitrary vector $v$ efficiently via the following recursive computation:
\begin{equation}
\label{eq:recurrence}
    C_{k+1}(s A + t I_{N})v = 2(s A + t I_{N})C_{k}(s A + t I_{N})v - C_{k-1}(s A + t I_{N})v,
\end{equation}
for $k = 1,...,m-1$, where $C_{0}(s A+t I_{N}) = I_{N}$, $C_{1}(s A+t I_{N}) = s A+t I_{N}$, and $I_{N}$ is the identify matrix of size $N$. The computation above only requires a fast algorithm for the matvec of $A$, which is available since $A$ is sparse. Algorithm \ref{algo:cheb-filter} is an example of Chebyshev polynomial filter which projects $[a_{0}, b]$ to $[-1,1]$ and $[a, a_{0})$ to $(-\infty, -1)$ respectively, where $a$ and $b$ are the lower and upper bounds of the spectrum respectively, and $a_{0}$ is the lower bound of the unwanted eigenvalues.
When the bounds $a$, $a_{0}$, and $b$ are unknown, they could be estimated by standard Lanczos decomposition via some matrix-vector products \cite{zhou2010block}. Note that $a$ and $b$ are $0$ and $2$ respectively when $A$ is the normalized Laplacian matrix of a graph, and in this case, $a_{0}$ could be roughly estimated, e.g., $a_{0} = a + (b-a)k_{want}/N$. Though the first estimation of $a_{0}$ might not be satisfactory, it will be approximated by Ritz values computed in Step 18 in Algorithm \ref{algo:bchdav-seq}.

\textbf{Orthonormalization (Step 6 of Algorithm \ref{algo:bchdav-seq}).} The orthonormalization step is initially performed by the DGKS
method \cite{daniel1976reorthogonalization}; random vectors are used to replace any vectors in $V(:, k_{sub}+1:k_{sub}+k_{b})$ that may become numerically linearly dependent to the current basis vectors in $V$. However, a parallel DGKS is inefficient and scales poorly to large concurrencies. We, therefore, replace parallel DGKS with the parallel TSQR \cite{demmel2012communication} for orthonormalization, which attains known communication lower bounds and communicates as little as
possible only up to polylogarithmic factors. We leave the description of the parallel TSQR in Section 3.3.

\textbf{Inner-outer restart (Steps 10 and 16  of Algorithm \ref{algo:bchdav-seq}).} When the dimension of the active space is larger than $act_{max}$ (Step 10, $k_{act}+k_{b} > act_{max}$), inner-restart would be applied to reduce the dimension of the active subspace to a smaller value $k_{ri}$. When the dimension of the current subspace is larger than $dim_{max}$ (Step 16, $k_{sub}+k_{b} > dim_{max}$), outer-restart would be used to reduce the dimension of the subspace to a smaller value $k_{ro}$. The default values of the optional parameters $act_{max}$, $k_{ro}$ and $k_{ri}$ can be readily determined by $k_{want}$, $k_{b}$ and
the matrix dimension; e.g.  $k_{ri} = max([\dfrac{act_{max}}{2}], act_{max}-3 k_{b})$ and $k_{ro} = dim_{max} - 2 k_{b} - k_{c}$.
Though the outer restart is a standard technique in eigensolvers, the inner restart is applied to reduce the cost of the orthonormalization in Step 6 and the Rayleigh-Ritz refinement in Steps 8, 9, and 11. The orthonormalization cost can be high since orthogonalizing $k$ vectors is of $O(Nk^2)$ complexity. A larger $dim_{max}$ may incur more orthonormalization cost at each iteration. In contrast, if we perform an inner restart, the number of non-converged vectors involved in the orthonormalization
per iteration is less than $act_{max}$. A Rayleigh-Ritz refinement includes computing the last column of the Rayleigh-quotient matrix $H$ (Step 8), computing an eigendecomposition of $H$ (Step 9), and refining the basis $V$ (Step 11). When $k_{want}$ is large and only standard outer restart is applied, the size of $H$ would become too large, making the last two steps expensive. Instead of waiting until the size of $H$ exceeds $dim_{max}$ to perform the standard restart, the algorithm performs an inner restart to restrict the active subspace $V(:,k_{c}+1:k_{c}+k_{act})$ and hence the size of $H$. The reduced refinement and reorthogonalization cost per iteration induced by the inner-outer restart may require more iterations to reach convergence; however, the total CPU time for the approach can be much smaller than that of only using standard restart \cite{zhou2010block}.

\textbf{Progressive filtering technique to utilize initial vectors (Step 17  of Algorithm \ref{algo:bchdav-seq}).} Instead of using subspace iteration on the entire available initial vectors at once, the algorithm progressively iterates over blocks of the initial vectors \cite{zhou2010block}. By assuming that the initial vectors are ordered so that their corresponding Rayleigh quotients are in non-decreasing order, the algorithm progressively filters all initial vectors in the natural order. First, it filters the first $k_{b}$ initial vectors when starting the iteration. Then, for each iteration afterward, it filters $e_{c}$ number of the leftmost unused initial vectors together with $k_{b}-e_{c}$ number of the current best non-converged Ritz vectors, then augment the $k_{b}$ filtered vectors into the projection basis. One advantage of this approach is that it can augment potentially better new basis vectors during the iteration process instead of only resorting to the initials \cite{zhou2010block}.

\section{Distributed Block Chebyshev-Davidson Algorithm}
With a basic understanding of the sequential algorithm, we are now ready to describe our distributed algorithm.
Algorithm \ref{algo:bchdav-dc} summarizes the framework of our distributed Block Chebyshev-Davidson method and Table \ref{tb:complexity} summarizes the complexity of each component of the distributed algorithm. To describe the algorithm and complexities in detail, we will progressively explain the following essential steps: distribution of matrices like $A$ and $V$, communication among processes, distributed SpMM, distributed Chebyshev polynomial filter, local matrix-matrix multiplication, and orthonormalization of tall-skinny matrices. 

{Before describing the algorithm, we want to highlight the scalability of the parallel algorithm. Though orthogonalization is used in the algorithm, it only takes a small portion in terms of computation costs due to the existence of the Chebyshev polynomial filter, which is scalable. See Figure \ref{fig:components-ratio} for the percentage of each algorithm component. Note that a higher degree of a Chebyshev polynomial filter results in faster convergence \cite{zhou2010block, zhou2007chebyshev} and more dominance among other components. Therefore, our parallel algorithm is more scalable than the eigensolvers, ARPACK, and LOBPCG, used in spectral clustering. See Figures \ref{fig:scalability-arpack-lobpcg} and \ref{fig:dbchdav-scaling} for demonstration.}

\begin{algorithm}
\caption{Distributed Block Chebyshev-Davidson method}
\label{algo:bchdav-dc}
\begin{algorithmic}[1]
\State Set up communicators of a 2D grid of $p$ processes and set $upperb$, $lowb$, and $low\_nwb$.
\State Set and distribute $V_{tmp} = V_{init}(:,1:k_{b})$, (construct $k_{b}-k_{init}$ random vectors if $k_{init} < k_{b}$); set $k_{i} = k_{b}$($k_{i}$ counts the number of used vectors in $V_{init}$).
\State Set $k_{c} = 0, k_{sub} = 0$ ($k_{sub}$ counts the dimension of the current subspace); set $k_{act} = 0$ ($k_{act}$ counts the dimension of the active subspace.)
\While{ $itmax$ is not exceeded}
\State $V(:,k_{sub}+1:k_{sub}+k_{b}) = DistributedChebyshevFilter(V_{tmp}, m, lowb, upperb, low\_nwb)$.
\State Orthonormalize $V(:,k_{sub}+1:k_{sub}+k_{b})$ against $V(:,1:k_{sub})$ using parallel TSQR.
\State Compute $W(:,k_{act}+1:k_{act}+k_{b}) = AV(:,k_{sub}+1:k_{sub}+k_{b})$ using distributed SpMM; set $k_{act}=k_{act}+k_{b}$; set $k_{sub} = k_{sub} + k_{b}$.
\State Parallelly compute the last $k_{b}$ columns of the Rayleigh-quotient matrix $H$: $H(1:k_{act},k_{act}-k_{b}+1:k_{act}) = V(:,k_{c}+1:k_{sub})^T W(:,k_{act}-k_{b}+1:k_{act})$, then symmetrize $H$.
\State Compute eigen-decomposition of $H(1:k_{act}, 1:k_{act})$ as $HY = YD$ locally, where $diag(D)$ is in non-increasing order. Set $k_{old} = k_{act}$.
\State If $k_{act} + k_{b} > act_{max}$, then do inner restart as: $k_{act} = k_{ri}, k_{sub} = k_{act} + k_{c}$.
\State Do subspace rotation locally (final step of Rayleigh-Ritz refinement) as: $V(:,k_{c}+1:k_{c}+k_{act}) = V(:,k_{c}+1:k_{old})Y(1:k_{old}, 1:k_{act}), W(:,1:k_{act}) = W(:,1:k_{old})Y(1:k_{old}, 1:k_{act})$.
\State Compute residual using distributed SpMM $r = AV(:,k_{c}+1: k_{c}+k_{b}) - V(:,k_{c}+1: k_{c}+k_{b})D(k_{c}+1: k_{c}+k_{b},k_{c}+1:  k_{c}+k_{b})$ and test for convergence of the $k_{b}$ vectors in $V(:,k_{c}+1, k_{c}+k_{b})$, denote the number of newly converged Ritz pairs at this step as $e_{c}$. If $e_{c} > 0$, then update $k_{c} = k_{c} + e_{c}$, save converged Ritz values in $eval$(sort $eval(:)$ in non-increasing order), and deflate/lock converged Ritz vectors (only need to sort $V(:,1:k_{c})$ according to $eval(1:k_{c})$).
\State If $k_{c} \geq k_{want}$, then return $eval(1:k_{c})$ and $V(:,1:k_{c})$, exit.
\State If $e_{c} > 0$, set $W(:,1:k_{act}-e_{c}) = W(:,e_{c}+1:k_{act}), k_{act} = k_{act} - e_{c}$.
\State Update $H$ as the diagonal matrix containing non-converged Ritz values $H = D(e_{c}+1:e_{c}+k_{act}, e_{c}+1:e_{c}+k_{act})$.
\State If $k_{sub} + k_{b} > dim_{max}$, do outer restart as: $k_{sub} = k_{c} + k_{ro}, k_{act} = k_{ro}$.
\State Get new vectors for the next filtering: Set $V_{tmp} = [V_{init}(k_{i}+1:k_{i}+e_{c}), V(:,k_{c}+1:k_{c}+k_{b}-e_{c})]; k_{i} = k_{i} + e_{c}$.
\State Set $low\_nwb$ as the median of non-converged Ritz value in $D$.
\EndWhile
\end{algorithmic}

\end{algorithm}

We assume that $p$ processes partitioning in parallel computation are organized into either a 2D $\sqrt{p}\times \sqrt{p} =p$ grid 
or a 1D grid. We use $P(i,j)$ where $0 \leq i,j < \sqrt{p}$ or $P(\ell)$ where $0\leq \ell < p$ to indicate a process at the corresponding location, respectively. For the former, the set of processes at row $i$ and column $j$ are respectively indicated with $P(i, :)$ and $P(:, j)$. Note that $P(i,j)$ and $P(j\sqrt{p} +i)$ refer to the same process. A matrix $M$ is partitioned in a 2D $\sqrt{p} \times \sqrt{p}$ block structure or a 1D $p$ row block structure. For the former, $M[i,j]$ denotes the submatrix associated with the $P(i,j)$ process, and for the latter, $M[i\sqrt{p} +j]$ denotes the row block associated with $P(i\sqrt{p} +j)$ or $P(j\sqrt{p} +i)$ depending on the context.

Collective communications play an important role in the algorithm. For communication cost analysis, we assume the cost of sending a message of size $w$ from one process to another is given by $\alpha + \beta w$, where $\alpha$ is the latency or message setup time, and $\beta$ is the reciprocal bandwidth or per-word transfer time. Five collective communication operations are applied to our work: MPI\_Bcast, MPI\_Reduce, MPI\_Allreduce, MPI\_Allgather, and MPI\_Reduce\_scatter.
If implemented with a tree algorithm, the MPI\_Bcast collective takes $O(\alpha \log p + \beta w \log p)$ cost to broadcast $w$ words to all processes in a communicator. The MPI\_Reduce collective reduces $w$ words from all processes at a single process, which has $O(\alpha \log p + \beta w \log p)$ cost with a tree implementation. The MPI\_Allreduce collective combines MPI\_Reduce and MPI\_Bast, which consequently has $O(\alpha \log p + \beta w \log p)$ cost as well. The MPI\_Allgather collective gathers $w$ words from all processes at each process, which has $O(\alpha \log p + \beta w p)$ cost if implemented with a recursive doubling algorithm. The MPI\_Reduce\_scatter collective collects $w$ words from each process and then scatters $w/p$ words to each process, which has $O(\alpha \log p + \beta w)$ cost assuming a recursive halving implementation. For details of these algorithms, refer to a survey by Chan et al. \cite{chan2007collective}.

\begin{figure}
\centering
\begin{tikzpicture}
  \node at (0.5,-0.5) {$U$};
  \node at (3.5,-0.5) {$A$};
  \node at (6.5,-0.5) {$V$};
  \node at (1.5,1.5) {=};
  \node at (5.5,1.5) {$\times$};
  \draw[help lines] (0,0) grid (1, 3);
  \draw (0,0.33) -- (1,0.33);
  \draw (0,0.67) -- (1,0.67);
  \draw (0,1.33) -- (1,1.33);
  \draw (0,1.67) -- (1,1.67);
  \draw (0,2.33) -- (1,2.33);
  \draw (0,2.67) -- (1,2.67);
  \node at (0.5, 0.5) {$U[7]$};
  
  \draw[help lines] (2,0) grid (5,3);
  \node at (3.5,0.5) {$A[2,1]$};
  
  \draw[help lines] (6,0) grid (7, 3);
  \draw (6,0.33) -- (7,0.33);
  \draw (6,0.67) -- (7,0.67);
  \draw (6,1.33) -- (7,1.33);
  \draw (6,1.67) -- (7,1.67);
  \draw (6,2.33) -- (7,2.33);
  \draw (6,2.67) -- (7,2.67);
  \node at (6.5, 1.166) {$V[5]$};
\end{tikzpicture}
\caption{\textbf{Illustration of A-Stationary 1.5D SpMM $U = AV$ when the number of processes is $p = 9$. Process $P(2,1)$ owns the submatrices $U[7], V[5]$, and $A[2,1]$.} }
\label{fig:spmm}
\end{figure}
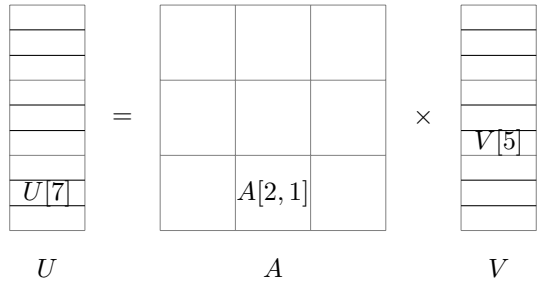

\subsection{Distributed SpMM}
Sparse times tall-skinny matrix multiplication (SpMM) is computing
\begin{equation}
\label{eq:spmm}
    U = AV
\end{equation}
where $V, U \in \mathbb{R}^{N\times k_{b}}$ and $k_{b} << N$. Such computations widely appear in Algorithm \ref{algo:bchdav-dc} including Steps 5, 7, and 12, so a fast distributed SpMM is of essential importance for distributed data. 

In the A-Stationary 1.5D algorithm \cite{selvitopi2021distributed}, $A$ is partitioned in 2D such that $P(i,j)$ process has $A[i,j]$; $V$ and $U$ are partitioned in 1D (by rows) such that $P(i,j)$ process has $V[j\sqrt{p}+i]$ and $U[i\sqrt{p}+j]$. See Figure \ref{fig:spmm} for illustration. $V$ is first replicated among $\sqrt{p}$ processes in each column of the process grid, and after this operation, $P(i, j)$ has $\sqrt{p}$ blocks of $V$, which are given by $V[j\sqrt{p} + \ell]$ for $0 \leq \ell < \sqrt{p}$. Next, the processes perform local SpMM of form $U^{j}[i\sqrt{p}+\ell] = A[i,j]B[j\sqrt{p}+\ell]$ for $0 \leq \ell < \sqrt{p}$, where $U^{j}$ denotes the partial dense resulting matrix evaluated by the process at $j-th$ column of the grid. These partial dense matrices are then summed up to get the final result matrix at each process with $U[i\sqrt{p} + j] = \Sigma_{\ell=0}^{\sqrt{p}} U^{\ell}[i\sqrt{p} + j]$. The memory requirement for each process is $nnz(A)/p + 2Nk_{b}/\sqrt{p}$ where $nnz(A)$ denotes the number of nonzeros in $A$, and the local computation takes $O((nnz(A)/p)k_{b})$ flops.

We use an MPI\_Allgather collective to realize the replication among $\sqrt{p}$ processes in each column of the process grid. The replication begins at each process with $Nk_{b}/p$ words, and thus the respective MPI\_Allgather has a cost of $O(\alpha \log p + \beta Nk_{b}/\sqrt{p})$. We use an MPI\_Reduce\_scatter collective with a summation operator to realize the reduction of partial dense matrices among $\sqrt{p}$ processes in each row of the process grid. The reduction begins at each process with $Nk_{b}/\sqrt{p}$ words, and thus the respective MPI\_Reduce\_scatter has a cost of $O(\alpha \log p + \beta Nk_{b}/\sqrt{p})$. Therefore, the total communication cost is 
\begin{equation}
\label{eq:communicationcost-spmm}
O(\alpha \log p + \beta \dfrac{2Nk_{b}}{\sqrt{p}}).
\end{equation}

In our distributed implementation Algorithm \ref{algo:bchdav-dc}, $A$ is partitioned in 2D, and $V, V_{init}, V_{tmp}, W$ are partitioned in 1D (by rows) in the same way. Each process maintains identical copies of other small dense matrices $D, Y, H$. The distributed SpMM is applied to Steps 5, 7, and 12. Applying Chebyshev polynomial filters in Step 5 requires additional treatments other than direct A-Stationary 1.5D.

In the distributed implementation in PARSEC, $A, V, V_{init}, V_{tmp}, W$ are all partitioned in 1D (by rows) in the same way, and then a straightforward 1D SpMM is applied. Though local computation in the 1D SpMM takes the same amount of flops, the communication cost 
\begin{equation}
\label{eq:communicationcost-spmm-parsec}
O(\alpha \log p + \beta Nk_{b})
\end{equation}
is much more expensive and not scalable because MPI\_Allgather is used for all $p$ processes simultaneously.

\subsection{Distributed Chebyshev Polynomial Filter}
To apply a polynomial of degree $m$ of a distributed sparse $A$ to a distributed dense matrix $V$, we must evaluate the distributed SpMM $AV$ $m$ times. Imagine applying a simple polynomial filter $x^{m}$ with $m = 2$:
\begin{equation}
\label{eq:AAV}
U_{2} = AAV, \quad V \in \mathbb{R}^{N\times k_{b}},
\end{equation}
we have to first compute $U_{1} = AV$ and then the filtering matrix $U_{2} = AU_{1}$.
In the A-Stationary 1.5D algorithm, $U_{1}$ and $V$ are partitioned in 1D but in different ways: the $P(i,j)$ process owns $V[j\sqrt{p}+i]$ and $U_{1}[i\sqrt{p}+j]$. Computing $AU_{1}$ directly via the A-Stationary 1.5D algorithm leads to the wrong filtering matrix when $A$ is a general sparse matrix. When $A$ is symmetric, $P(j,i)$ owns $U_{1}[j\sqrt{p}+i]$ and the submatrices owned by $P(i,j)$ and $P(j,i)$ are the same. As a consequence, transposing the process grid and then applying the A-Stationary 1.5D algorithm to $A$ and $U_{1}$ gives the correct filtering matrix $U_{2}$ because the operations are equal to computing $A^{T}U_{1}$. Note that $V$ and $U_{2}$ need not be two separate variables in implementation. However, this only works for the case when $m$ is even. When $m$ is odd, the final filtering matrix $U_{m}$ after applying a polynomial of degree $m$ to $V$ is still partitioned differently (like $U_{1}$). There are two remedies.

\begin{itemize}
\item a) When $m$ is odd, we partition an identity matrix $I$ in the same way as we partition $A$, transpose the process grid, and then apply A-Stationary 1.5D to $I$ and the filtering matrix $U_{m}$ to get the final matrix which is partitioned in the same way as $V$;
\item b) After each distributed SpMM, we apply a) to the resulting matrix to get a re-distributed matrix partitioned in the same way as $V$. It is clear that this remedy is more expensive than the first one, but the total complexities are still in the same order.
\end{itemize}
Note that applying a distributed SpMM to the identity matrix $I$ and a matrix $U$ is essentially equal to moving around parts of the matrix $U$ without any local computations. Our distributed Chebyshev polynomial filter of degree $m$ is implemented in the second way because every intermediate matrix $U_{i}(i <= m)$ and $V$ should be partitioned in the same way due to the addition operations in Step 5 and 8 in Algorithm \ref{algo:cheb-filter}. Algorithm \ref{algo:cheb-filter-dc} summarizes the distributed Chebyshev polynomial filter algorithm. The local computation takes $O(nnz(A)mk_{b}/p)$ because each process averagely own $nnz(A)/p$ nonzeros of $A$,  and the total communication cost is 
\begin{equation}
\label{eq:communicationcost-filter}
O(m\alpha \log p + \beta \dfrac{2mNk_{b}}{\sqrt{p}}).
\end{equation}

Despite applying the 1D SpMM in PARSEC to compute the filtering matrices is straightforward, which avoids the issues mentioned above, it requires much more communication cost
\begin{equation}
\label{eq:communicationcost-filter-parsec}
O(m\alpha \log p + \beta mNk_{b}),
\end{equation}
which is not scalable.

\begin{algorithm}
\caption{$[W] = DistributedChebyshevFilter(V, m, a, b, a_{0})$}
\label{algo:cheb-filter-dc}
\begin{algorithmic}[1]
\State \textbf{variables}: $\Omega$ the 2D grid of $p$ processes; $A$ the distributed sparse matrix; $I$ the distributed identity matrix; $V$ the distributed input matrix; $m$ the degree of a Chebyshev polynomial; $a$ the lower bound of unwanted eigenvalues of $A$; $b$ the upper bound of all eigenvalues; $a_{0}$ the lower bound of all eigenvalues; 

\State $c = (a + b) / 2$; $e = (b - a)/2$;
\State $\sigma = e/(a_{0}-c)$;
\State $t = 2/\sigma$;
\State Compute $U = (AV - cV)\sigma/e $ using distributed SpMM;
\State Transpose the 2D grid $\Omega$;
\State Update $U \hookleftarrow I U$ using distributed SpMM;
\For{$i = 2:m$}
\State $\sigma_{1} = 1/(\tau - \sigma)$;
\State Transpose the 2D grid $\Omega$;
\State Compute $W = 2\sigma_{1}(A U - c U)/e - \sigma\sigma_{1}v$ using distributed SpMM;
\State Transpose the 2D grid $\Omega$;
\State Update $W \hookleftarrow I W$ using distributed SpMM;
\State $V = U$;
\State $U = W$;
\State $\sigma = \sigma_{1}$;
\EndFor
\end{algorithmic}

\end{algorithm}

\subsection{Distributed Orthonormalization}
For orthonormalization, we adopt the parallel TSQR in \cite{demmel2012communication}. We begin by illustrating the algorithm for the case of $p = 4$ processes and then state the general version of the algorithm and its performance model. Suppose that the $N \times n$ matrix $V$ is divided into four row blocks $V = [V[0];V[1];V[2];V[3]]$, where $V[i]$ is $N/4\times n$ associated with $P(i)$ process. First, process $i$ computes the QR factorization of its row block $V[i] = Q_{i}R_{i}$, see eq. (\ref{eq:tsqr}).
\begin{equation}
\label{eq:tsqr}
V= \begin{pmatrix}
Q_{0}R_{0} \\
Q_{1}R_{1} \\
Q_{2}R_{2} \\
Q_{3}R_{3} 
\end{pmatrix},
\begin{pmatrix}
\begin{pmatrix}
R_{0} \\
R_{1} \\
\end{pmatrix}\\
\begin{pmatrix}
R_{2} \\
R_{3} \\
\end{pmatrix}
\end{pmatrix}
= 
\begin{pmatrix}
Q_{01}R_{01} \\
Q_{23}R_{23} \\
\end{pmatrix},
\begin{pmatrix}
R_{01} \\
R_{23} \\
\end{pmatrix} = Q_{0123}R_{0123}.
\end{equation}
Then, processes work in pairs, combining their local $R$ factors by computing the QR factorization of the $2n\times n$ matrix $[R_0;R_1]$, $[R_2; R_3]$, respectively. Thus, $[R_0; R_1]$ is replaced by $R_{01}$ and $[R_2; R_3]$ is replaced by $R_{23}$. Here and later, the subscripts on a matrix like $R_{ij}$ refer to the original row blocks $V[i]$ and $V[j]$ on which they depend. Finally, the $2n\times n$ QR factorization $[R_{01};R_{23}] = Q_{0123}R_{0123}$ is computed. It is easy to verify that $R_{0123}$ is the $R$ factor in the QR factorization of the original matrix $V$. We could combine all the steps above into eq. (\ref{eq:tsqrequi}), which expresses the entire computation as a product of intermediate orthonormal factors. Note that the dimensions of the intermediate Q factors are chosen consistently for the product to make sense. The Q factor of the QR decomposition is then the product of the intermediate orthonormal factors. Note that the Q factor is computed locally without communication cost because each process owns all the corresponding intermediate orthonormal factors. 
% By uniqueness of the QR decomposition, this is the QR decomposition of $V$.

\begin{equation}
\label{eq:tsqrequi}
V = 
\begin{pmatrix}
Q_{0} & & & \\
& Q_{1} & & \\
& & Q_{2} & \\
& & & Q_{3} \\
\end{pmatrix}
\begin{pmatrix}
Q_{01} & \\
& Q_{23} \\
\end{pmatrix}
Q_{0123}R_{0123}.
\end{equation}

The parallel TSQR for general cases is summarized in Algorithm \ref{algo:paralleltsqr} (Algorithm 1 in \cite{demmel2012communication}). In the implementation, the communication between process $i$ and its $q-1$ neighbor in Step 5, 6, and 7 are achieved by MPI\_Allgather, which take $O(\alpha \log q + \beta n^2 q)$ time each level. Note that $q$ is a constant. The local QR decomposition at the leaf level and a non-leaf level take $O(2N n^2/p - 2 n^3 / 3)$ and $O(2qn^3 - 2n^3/3)$ flops, respectively. The local update of $Q$ at the leaf level and a non-leaf level take $O(Nn^2/p)$ and $O(qn^3)$, respectively. When orthonormalizing $V$ in Step 6 of Algorithm \ref{algo:bchdav-dc}, the second dimension $n$ of $V$ is bounded by $act_{max}$. Therefore, in our distributed orthonormalization with TSQR, it takes totally
\begin{equation}
   O(3Nact_{max}^2/p + 3act_{max}^3 \log p) 
\end{equation}
flops for local computation and 
\begin{equation}
O(\alpha \log p + \beta act_{max}^2 \log p)    
\end{equation}
time for communication.

Unlike the parallel TSQR, to orthonormalize $V(:,k_{sub}+1:k_{sub}+k_{b})$ against $V(:,1:k_{sub})$ in Step 6 of Algorithm \ref{algo:bchdav-dc}, the parallel DGKS first orthonormalizes $V(:,k_{sub}+1)$ against $V(:,1:k_{sub})$, then orthonormalizes $V(:,k_{sub}+2)$ against $V(:,1:k_{sub}+1)$, and repeats the procedures until all vectors are orthonormal. Despite simple, the local computation takes $O(Nact_{max}k_b^2 / p)$ flops and communication takes 
\begin{equation}
O(\alpha k_{b} \log p + \beta \dfrac{N k_b}{p} \log p)    
\end{equation}
time because MPI\_Allreduce is applied to normalization. Note that the communication cost is much higher than that of the parallel TSQR because $Nk_b / p$ is much larger than the constant $act_{max}^2$ when the problem dimension $N$ is large. 

\begin{algorithm}
\caption{parallel TSQR}
\label{algo:paralleltsqr}
\begin{algorithmic}[1]
\State \textbf{Require}: set $\Pi$ of $p$ processes; tree with $p$ leaves and height $L = \log_{q} p$, describing communication pattern; $N\times n$ matrix $V$ distributed in 1D; current process index $i$.
\State Compute QR factorization $A[i] = Q_{i,0}R_{i,0}$
\For{$k = 1:L$}
\If{the current process $i$ has $q-1$ neighbors}
\State Send $R_{i,k-1}$ to each neighbor 
\State Receive $R_{j,k-1}$ from each neighbor j
\State Stack the $R_{j,k-1}$ from all neighbors (incl. $R_{i,k-1}$), in $j$ order, into the $q n \times n$ matrix $C_{i,k}$ , and factor $C_{i,k} = Q_{i,k} R_{i,k}$
\Else
\State $R_{i,k} := R_{i,k-1}$, and $Q_{i,k} := I_{n\times n}$
\EndIf
\EndFor
\State \textbf{Ensure}: $R_{i,L}$ is the R factor of A, for all processes.
\State \textbf{Ensure}: Q factor could be evaluated locally and top-down using the tree of intermediate $Q$ factors
${Q_{i,k}: i \in \Pi, k \in {0,1,...,L}}$.
\end{algorithmic}

\end{algorithm}

\subsection{Other Steps}
In our algorithm, each process maintains the same Rayleigh-quotient Matrix $H$ which is updated in Step 8. To update $H(1:k_{act}, k_{act}-k_{b}+1:k_{act})$, we first perform the local computation of the transpose of the corresponding parts of $V[j\sqrt{p}+i]$ and $W[j\sqrt{p}+i]$ on process $P(i,j)$, then reduce all results among $\sqrt{p}$ processes in each row by MPI\_Allreduce with summation operator. Next, we perform a similar reduction for all resulting among $\sqrt{p}$ processes in each column to get the updated $H$. This step takes at most $O(Nk_{b}act_{max}/p)$ flops for local computation and 
\begin{equation}
\label{eq:communicationcost-H}
O(\alpha \log p + \beta act_{max}k_{b} \log p)
\end{equation}
for communication, where $act_{max}$ and $k_{b}$ are small constants.

Each process also maintains the same copy of $D$ and $Y$ at each process because they are of dimension at most $act_{max}$. Computing residual in Step 12 involves a distributed SpMM, a local matrix-matrix multiplication, and communication, which in total takes $O(nnz(A)k_{b}/p + Nk_{b}^2/p)$ flops for local computation and 
\begin{equation}
\label{eq:communicationcost-residual}
O(\alpha \log p + \beta \dfrac{2Nk_{b}}{\sqrt{p}}).
\end{equation}
time for communication.

Steps 9, 11, and 15 are performed locally without any communication cost and together take at most $O(act_{max}^3 + N act_{max}^2/p)$ flops. All other steps take only $O(1)$ time.

\begin{figure}[ht!]
  \begin{center}
    \begin{tabular}{c}
      \includegraphics[scale=0.35]{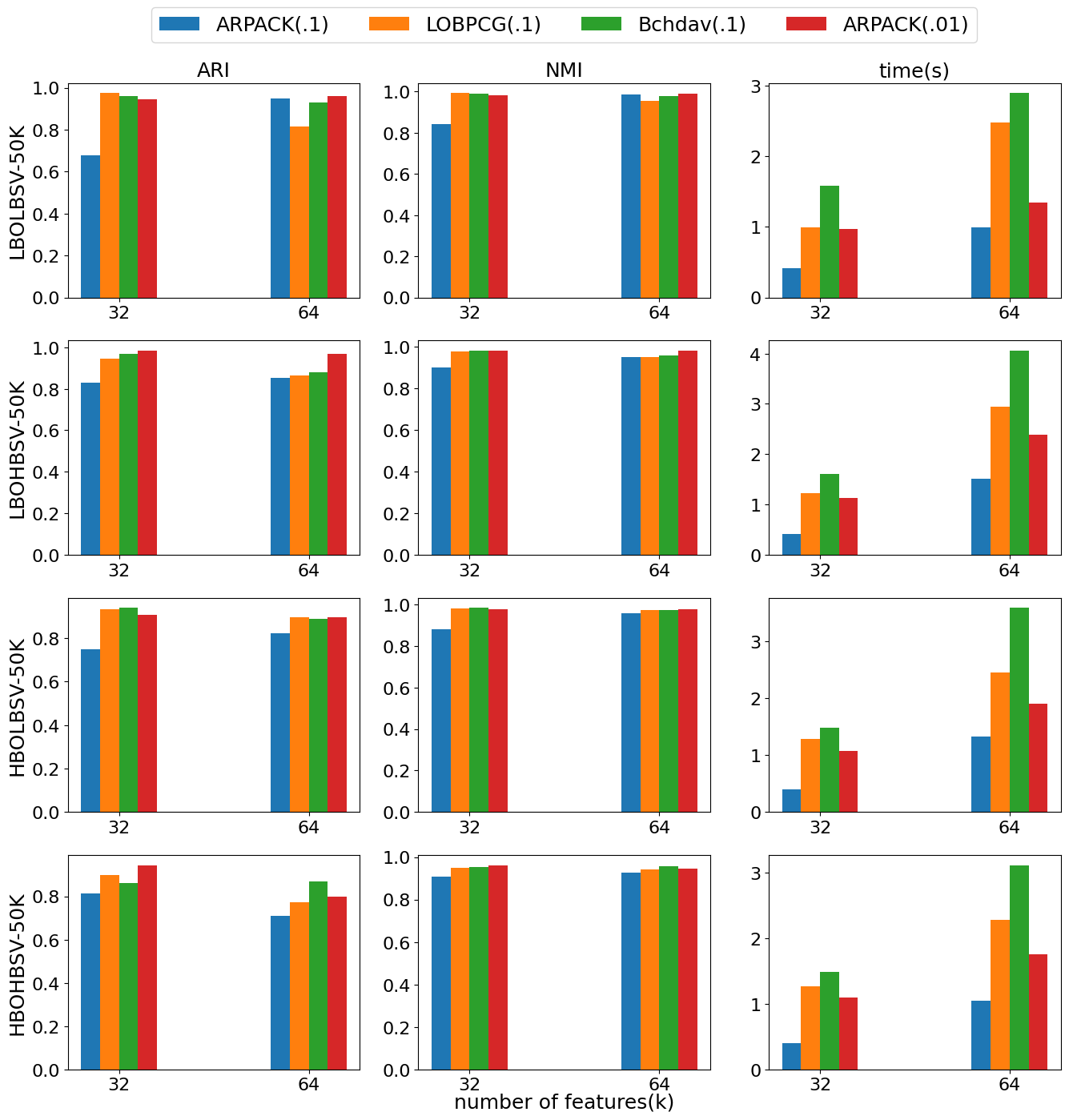} 
      
    \end{tabular}
  \end{center}
\caption{Comparisons of ARPACK, LOBPCG without preconditioning, and the Block Chebyshev-Davidson method (Bchdav) in clustering performance on graphs with 50 thousand nodes. ARPACK runs with tolerance $.1$ and $.01$. LOBPCG and Bchdav run with tolerance $.1$. In Bchdav, $k_b = 4$ and $m = 11$.}
\label{fig:comp50k}
\end{figure}

\begin{figure}[ht!]
  \begin{center}
    \begin{tabular}{c}
      \includegraphics[scale=0.35]{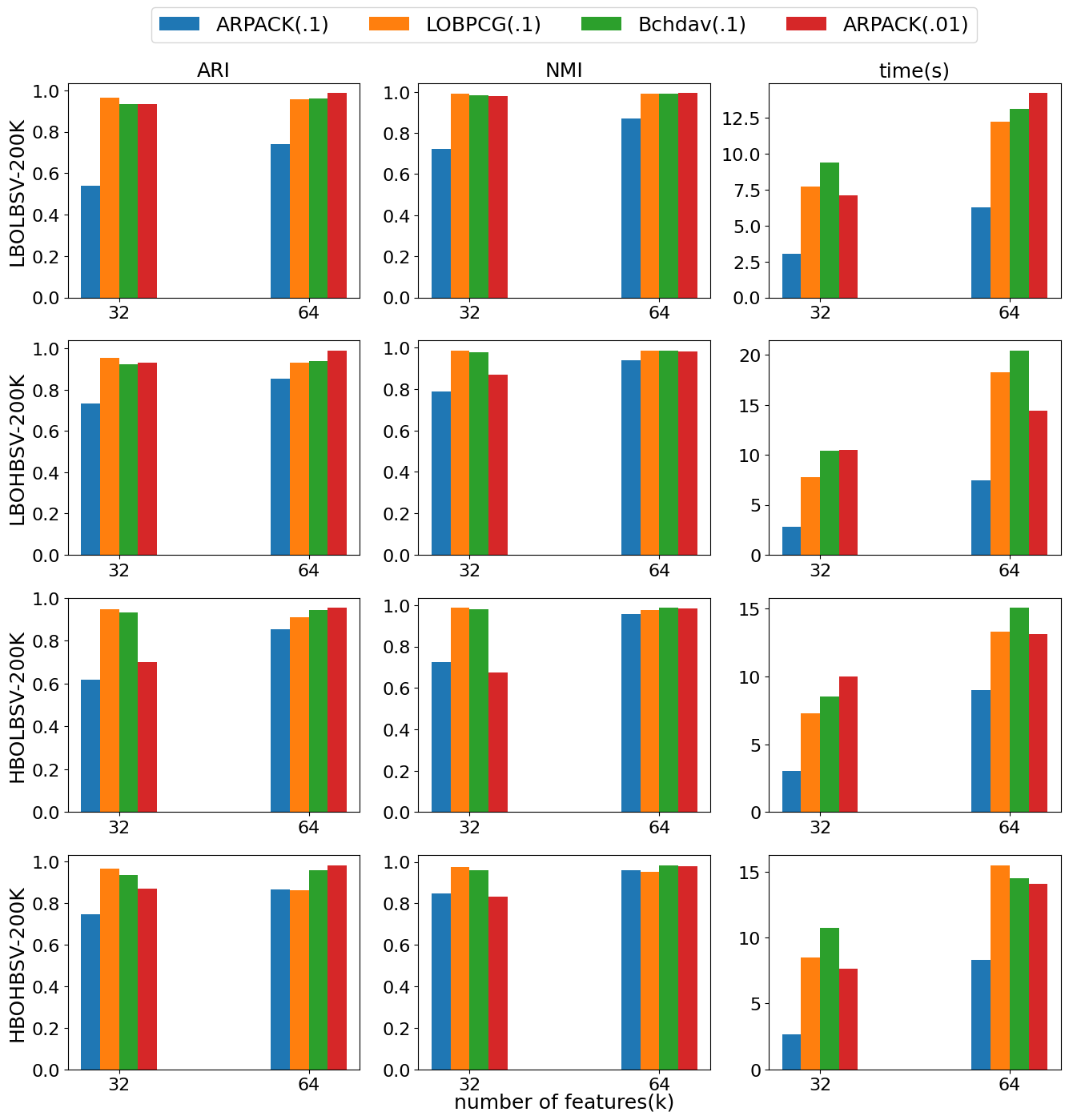} 
      
    \end{tabular}
  \end{center}
\caption{Comparisons of ARPACK, LOBPCG without preconditioning, and the Block Chebyshev-Davidson method (Bchdav) in clustering performance on graphs with 200 thousand nodes. ARPACK runs with tolerance $.1$ and $.01$. LOBPCG and Bchdav run with tolerance $.1$. In Bchdav, $k_b = 4$ and $m = 11$.}
\label{fig:comp200k}
\end{figure}

\begin{figure}[ht!]
  \begin{center}
    \begin{tabular}{c}
      \includegraphics[scale=0.35]{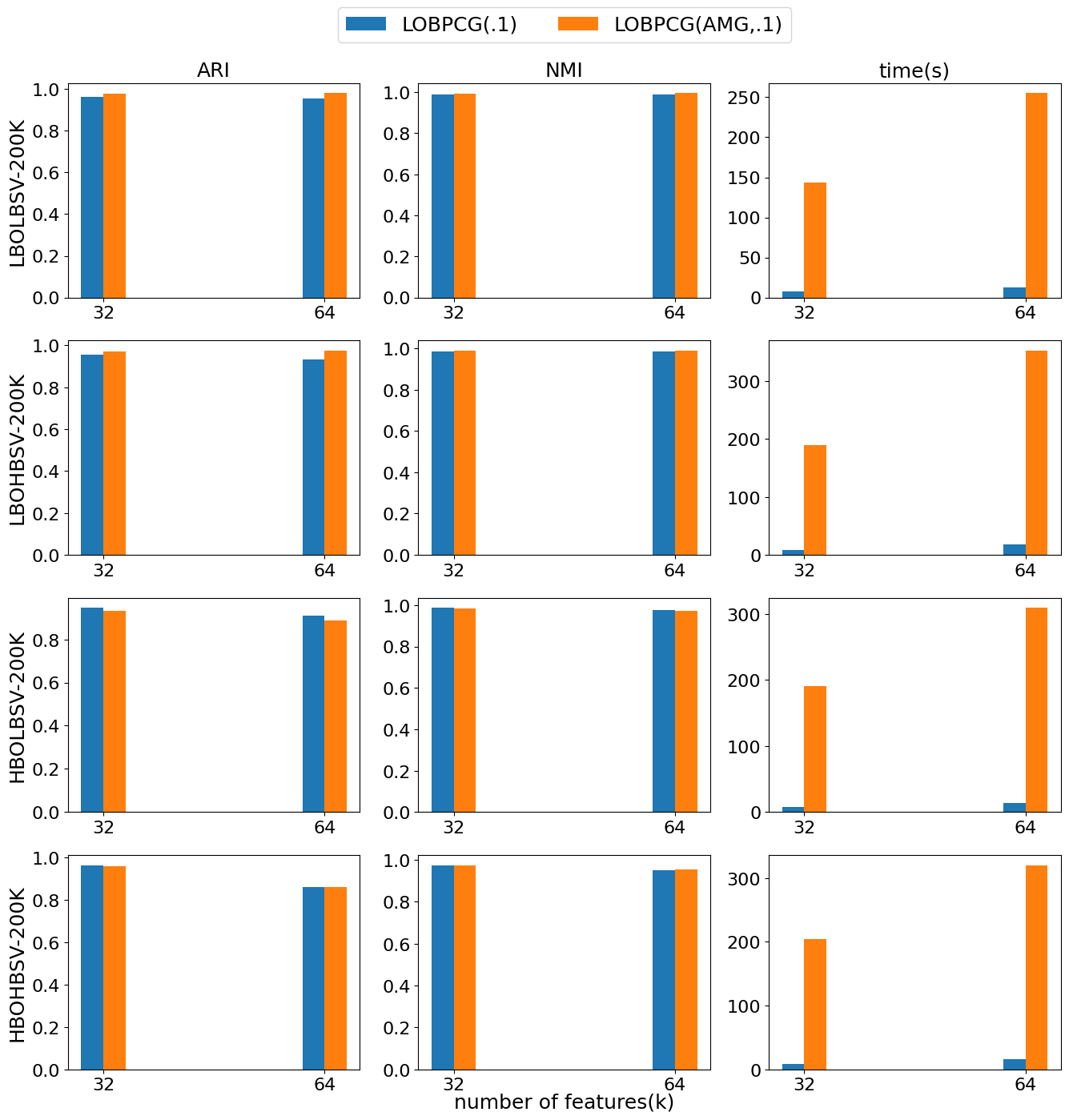} 
      
    \end{tabular}
  \end{center}
\caption{Comparisons of LOBPCG with and without preconditioning (AMG) in clustering performance on graphs with 200 thousand nodes. LOBPCG runs with tolerance $.1$.}
\label{fig:compamg}
\end{figure}

\section{Numerical Results}
{We present the numerical results of three experiments to demonstrate the effectiveness and advantages of our algorithm.
\begin{itemize}
    \item[] 1. In the first experiment, we apply our algorithm as the eigensolver in spectral clustering to partition graphs with known truth from the IEEE HPEC Graph Challenge \footnote{http://graphchallenge.mit.edu}. We compare the sequential version of our method with the eigensolvers, ARAPCK, and LOBPCG, used in spectral clustering. Numerical results demonstrate the effectiveness of our method when used for spectral clustering.
    \item[] 2. In the second experiment, we test the scalability of parallel ARPACK, parallel LOBPCG, and the parallel version of our method to show that our method is more scalable.
    \item[] 3. In the final experiment, we show the superiority of our parallel implementation to the parallel implementation of the Block Chebyshev-Davidson method in PARSEC.
\end{itemize}
  }

All the numerical experiments conducted on the Zaratan cluster operated by the University of Maryland. It features 360 compute nodes, each with dual 2.45 GHz AMD 7763 64-core CPUs. The cluster has HDR-100 (100 Gbit) Infiniband interconnects between the nodes, with storage and service nodes connected with full HDR (200 Gbit). The theoretical peak floating-point rate is 3.5 Pflops. The sequential version of our method is written in Matlab, and the parallel version is developed in Julia 1.7.3 using MPI.jl \cite{byrne2021mpi} with OpenMPI \cite{gabriel2004open}. The code\footnote{https://github.com/qiyuanpang/DistributedLEVP.jl} is available on GitHub. Standard variable settings used in the numerical experiments are as follows:
\begin{itemize}
    \item $p$: the number of processes or cores;
    \item $k$: the number of eigenvectors to compute;
    \item $N$: the dimension of a matrix or the number of nodes in a graph;
    \item $m$: the degree of a Chebyshev polynomial filter;
    \item $k_{b}$: the number of vectors added to the projection basis per iteration;
    \item $act_{max}$: the maximum dimension of the active subspace $act_{max} = max(5 k_{b}, 30)$ throughout all experiments;
    \item $dim_{max}$: the maximum dimension of the subspace $dim_{max} = max(act_{max}+2 k_{b}, k+30)$ throughout all experiments;
    \item $tol$: the stopping criterion or tolerance. 
\end{itemize}

\subsection{Comparisons of eigensolvers used in spectral clustering}
{The exact bounds of the Laplacian or normalized Laplacian $A$ are known, e.g., the smallest and largest eigenvalues of the normalized Laplacian are $0$ and $2$, respectively. The Block Chebyshev-Davidson method, therefore, converges fast with the optimal bounds. Here, we consider the symmetric normalized Laplacian $A$ of static undirected graphs from the IEEE HPEC Graph Challenge. There are four categories of graphs: low block overlap and low block size variation (LBOLBSV), low block overlap and high block size variation (LBOHBSV), high block overlap and low block size variation (HBOLBSV), and high block overlap and high block size variation (HBOHBSV). We use notations like "LBOLBSV-200K" to denote different graphs, where "LBOLBSV" indicates the graph category and "200K" indicates the graph owns 200 thousand nodes. For the symmetric normalized Laplacian of each graph, we apply eigensolvers to compute $k=32$ or $k=64$ eigenvectors and then apply K-means clustering to partition the graph. Since the true partitions are known for each graph, the number of clusters in K-means clustering is always set to be the number of true partitions. We adopt external indexes Adjusted Rand Index (ARI) \cite{hubert1985comparing} and Normalized Mutual Information (NMI) \cite{danon2005comparing} to measure clustering quality. For ARI and NMI, values close to 0 indicate that the two clusterings are primarily independent, while values close to 1 indicate significant agreement. ARI is adjusted against chance, while NMI is not. To alleviate the randomness in K-means clustering, we repeat each experiment $20$ times to record the average indexes. 

% For every type of graph, k-means clustering based on the computed eigenvectors reaches a high rand index which is one of the major metrics measuring clustering accuracy. Note that k-means is not guaranteed to find the optimal clusters \cite{hartigan1979algorithm} and the task is indeed NP-hard in general Euclidean space even for two clusters \cite{aloise2009np, dasgupta2009random}.

Figure \ref{fig:comp50k} and \ref{fig:comp200k} summarize the comparisons of sequential ARPACK, LOBPCG, and the Block Chebyshev-Davidson method when used for spectral clustering. LOBPCG is sometimes used with Algebraic multigrid (AMG) preconditioning in spectral clustering, but AMG preconditioning is not always effective \cite{knyazev2017recent}. Figure \ref{fig:compamg} shows that AMG preconditioning does not improve clustering quality on the graphs we considered but takes additional expensive costs. Hence, in this experiment, we only used LOBPCG without preconditioning. The numerical results show that the Block Chebyshev-Davidson method has the top clustering quality, though it is usually a bit slower than ARPACK and LOBPCG with $.1$ tolerance. With $.1$ tolerance, ARPACK achieves the worst clustering quality. Even with a $.01$ tolerance, ARPACK sometimes reaches worse clustering quality compared to LOBPCG and Block Chebyshev-Davidson method with $.1$ tolerance. To conclude, Block Chebyshev-Davidson achieves competitive clustering performance even though it is a bit more computationally expensive. However, in the next experiment, we will show that Block Chebyshev-Davidson is more scalable in parallel computing environments.}

% \begin{figure}[ht!]
%   \begin{center}
%     \begin{tabular}{c}
%       \includegraphics[height=4.0in]{results/compeigl_k16_deg11_blk4_tol-6.png}
%     \end{tabular}
%   \end{center}
% \caption{\textbf{CPU time comparison between chebdav and LOBPCG for converging $16$ eigenpairs of matrices in Tables \ref{tb:spectralclustering-1}: LBOLBSV (upper left), LBOHBSV (upper right), HBOLBSV (lower left), and HBOHBSV (lower right). The stopping criteria (tolerance) is $10^{-6}$. For chebdav, the degree $m = 11$ and $k_{b} = 4$. The preconditioner for LOBPCG is the identity matrix because no suitable preconditioners are known.}
% }
% \label{fig:eigensolvers-comparison}
% \end{figure}

% \subsection{Comparison of Sequential Eigensolvers}
% We compare the sequential Block Chevyshev-Davidson method with a block preconditioned eigensolver LOBPCG \cite{knyazev2001toward} on the Laplacian matrices in Tables \ref{tb:spectralclustering-1}. The preconditioner of LOBPCG is the identity matrix because we do not know any other effective preconditioners, and the design for preconditioners of LOBPCG is an entirely different topic. Figure \ref{fig:eigensolvers-comparison} presents that the Block Chebyshev-Davidson method is comparable to LOBPCG while computing eigenpairs of the normalized Laplacian matrices of the graphs. Furthermore, numerical results in \cite{zhou2010block} show that the Block Chebyshev-Davidson method is an order of magnitude faster than LOBPCG when computing the smallest eigenpairs for specific matrices from quantum chemistry.

\begin{table}[!htbp]
\caption{\textbf{Properties of matrices used in our evaluations. The values under the ``load imb." column present the load imbalance in terms of the sparse matrix elements for 121 processes (i.e., 11 $\times$ 11 2D partition).}}
\centering
\scalebox{1.0}{
\begin{tabular}{|c|c|c|c|c|}
\hline
Sparse Matrix & N & avg degree & nnz(A) & load imb.\\
\hline
LBOLBSV & 5M & 48.5 & 242M & 1.21 \\
HBOLBSV & 20M & 48.5 & 970M & 1.21 \\
MAWI-Graph-1 & 18M & 3.0 & 56M & 8.8 \\
Graph500-scale24-ef16 & 16M & 31.6 & 529M & 7.15 \\
\hline

\end{tabular}
}
\label{tb:matrixproperties}
\end{table}

\begin{figure}[ht!]
  \begin{center}
    \begin{tabular}{cc}
      \includegraphics[scale=0.35]{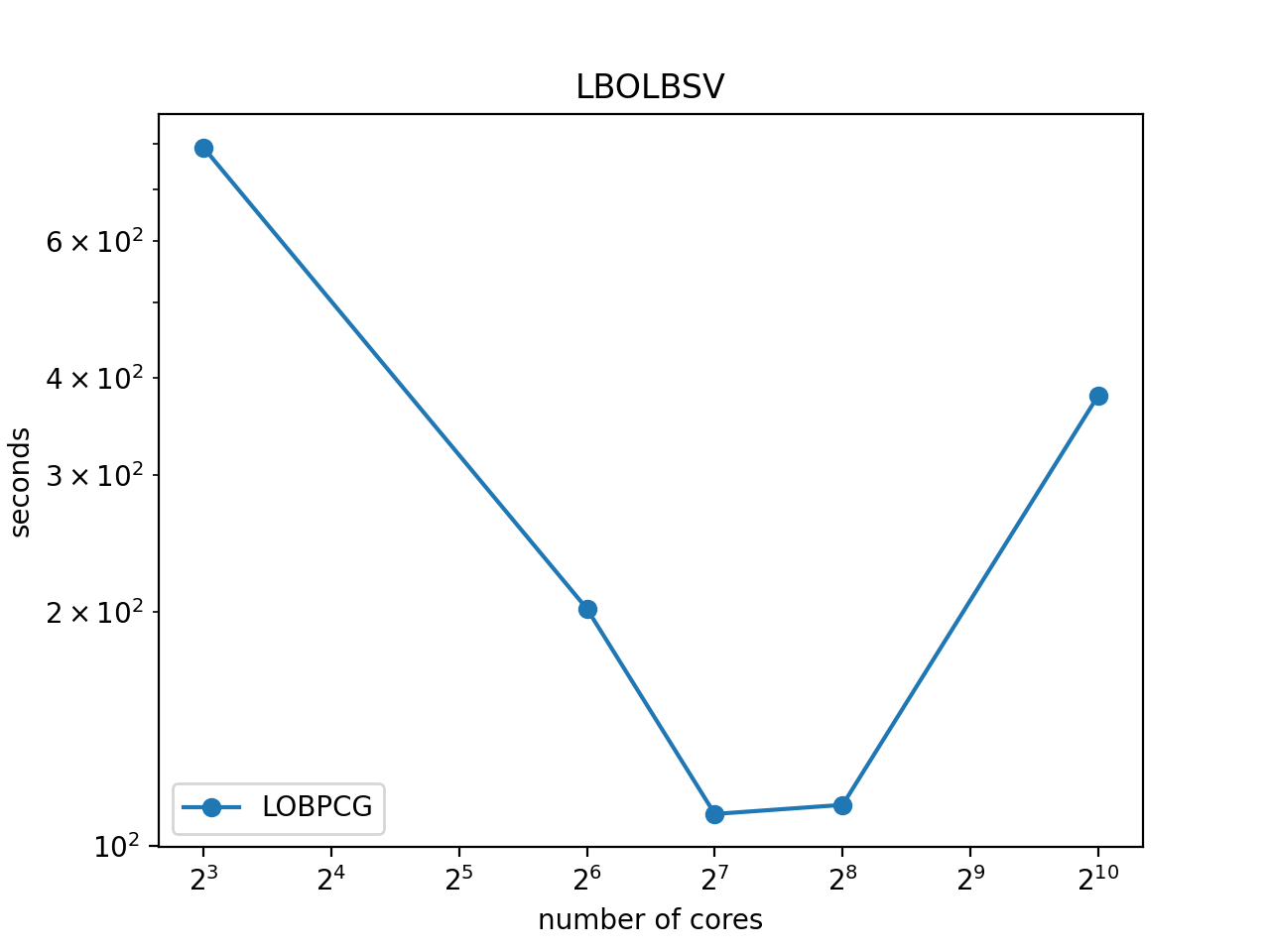} &
      \includegraphics[scale=0.35]{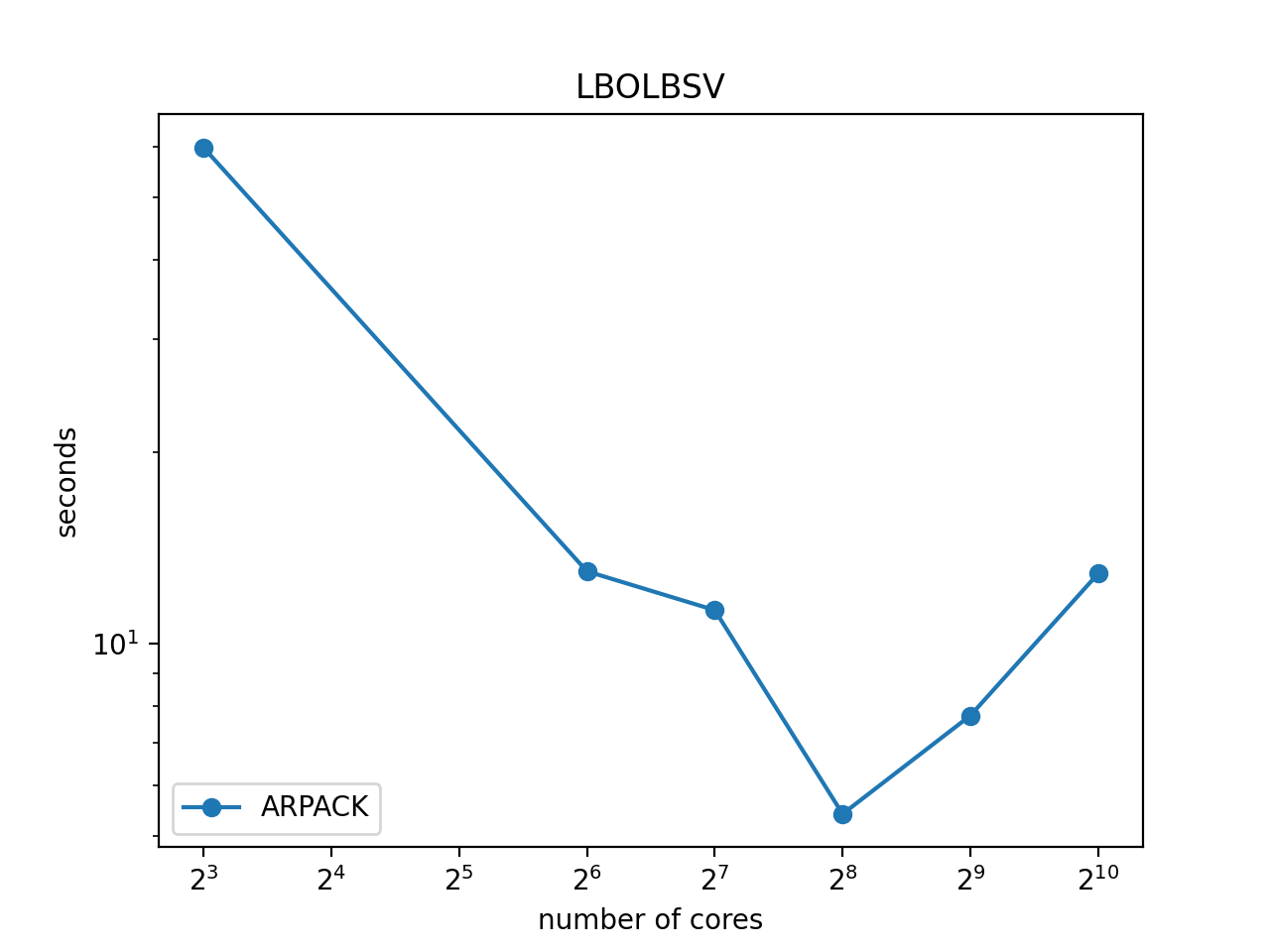} 
      
    \end{tabular}
  \end{center}
\caption{Scaling of parallel ARPACK and LOBPCG to compute $k = 64$ eigenvectors of the matrix LBOLBSV(SG)-1M with tolerance $0.01$. }
\label{fig:scalability-arpack-lobpcg}
\end{figure}

\begin{figure}[ht!]
  \begin{center}
    \begin{tabular}{ccc}
      \includegraphics[height=1.5in]{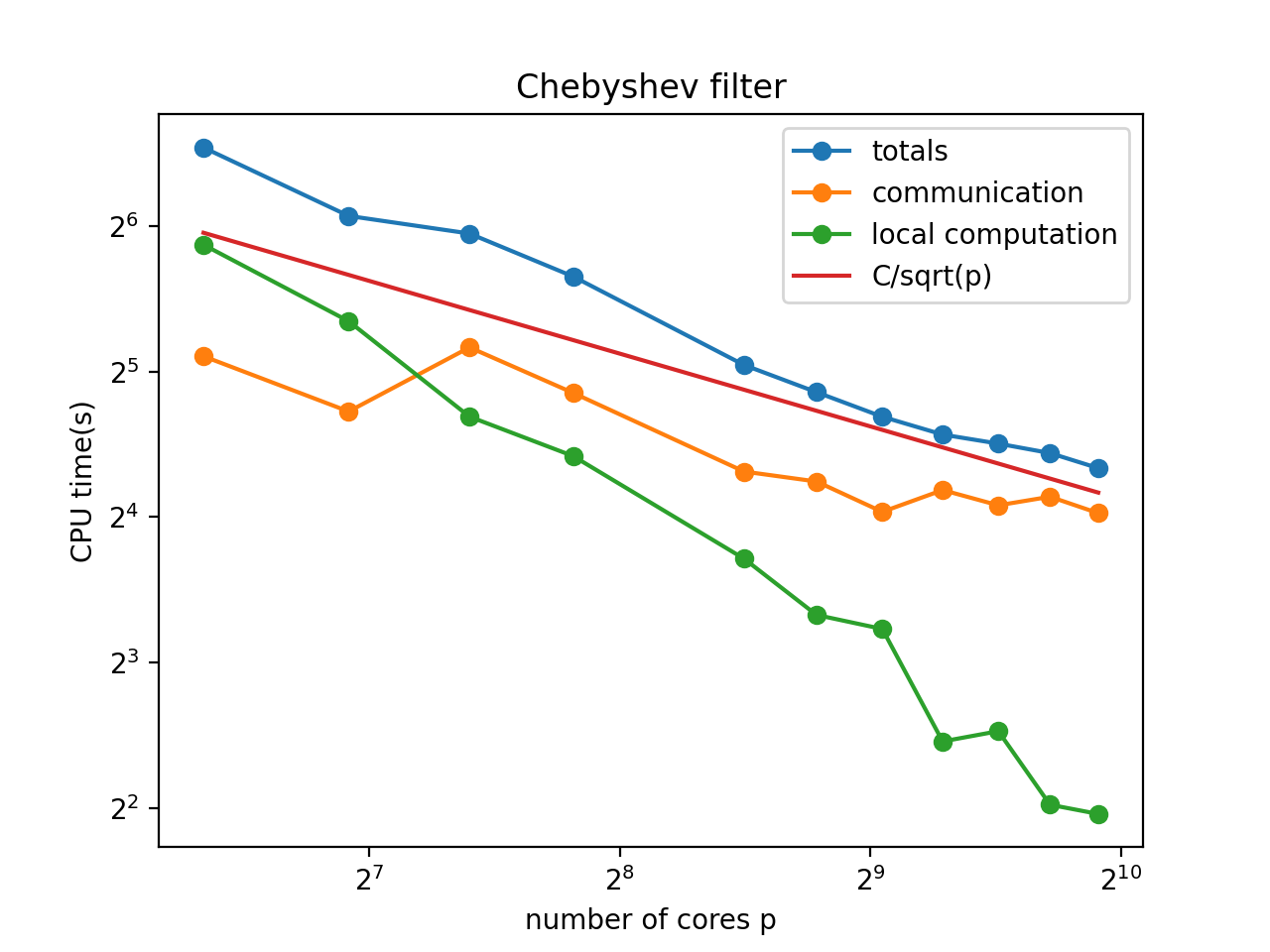} &
      \includegraphics[height=1.5in]{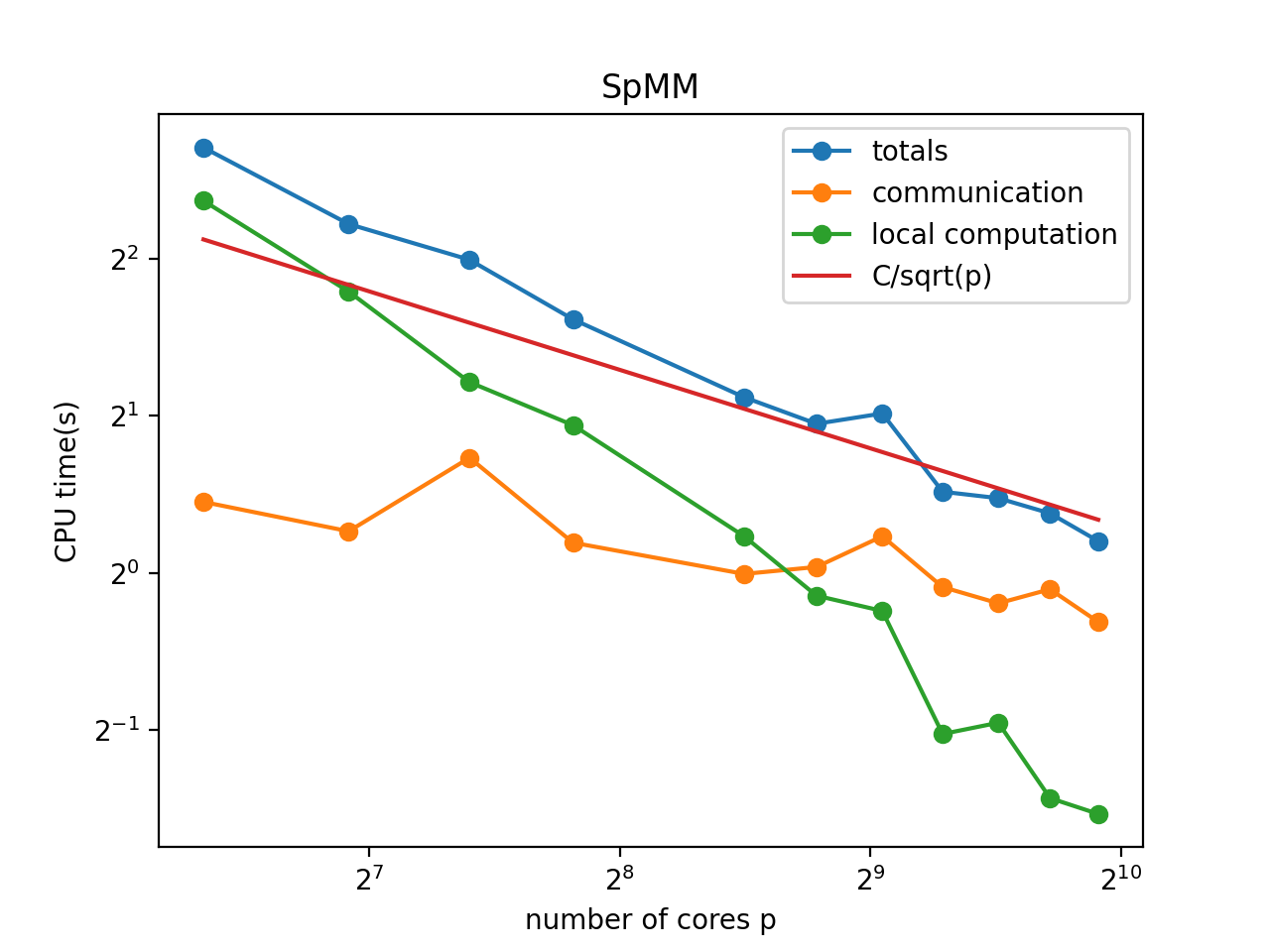} &
      \includegraphics[height=1.5in]{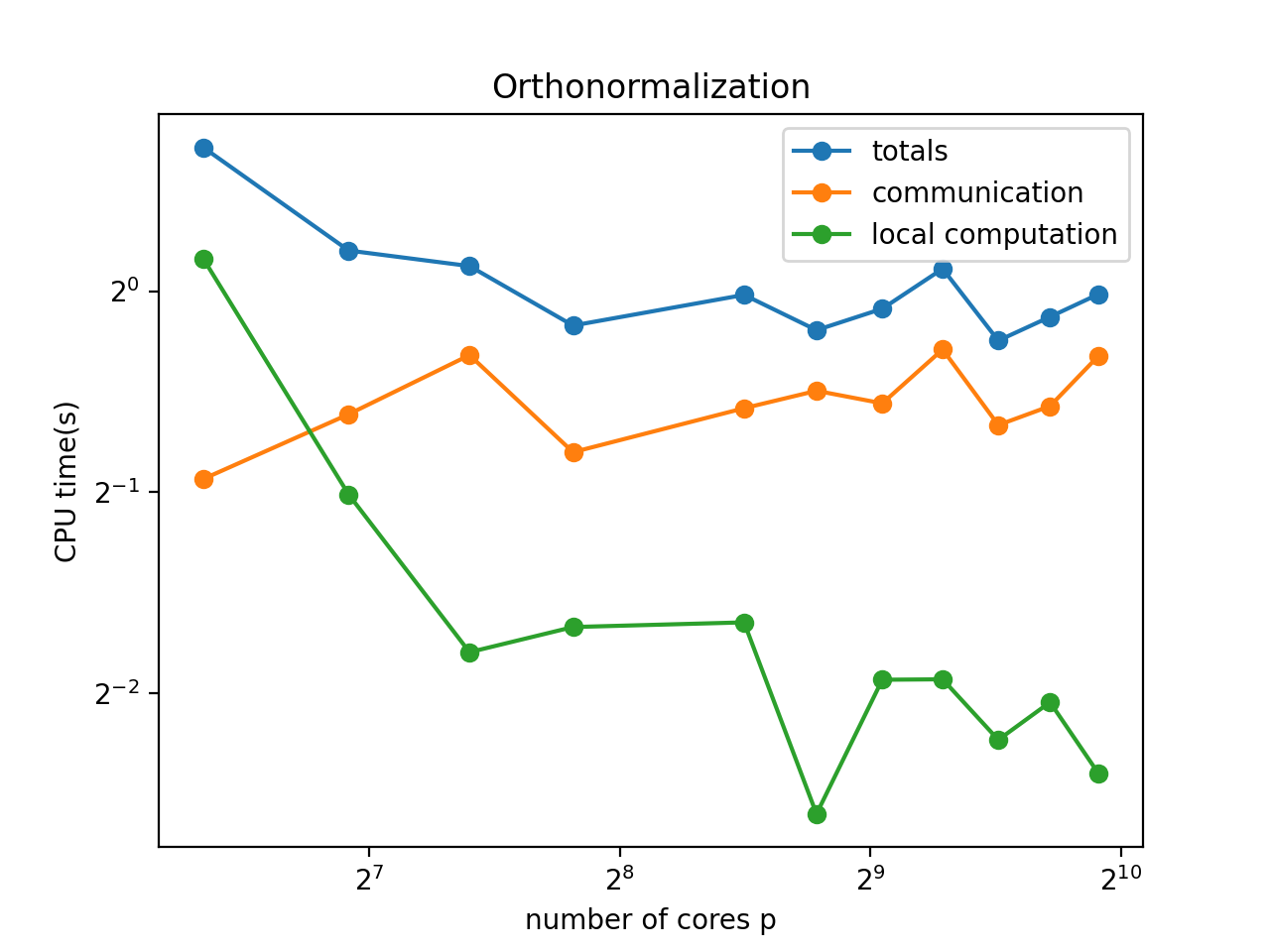} \\
    \end{tabular}
  \end{center}
\caption{\textbf{Scaling of local computation and communication in a distributed Chebyshev filter, SpMM, and TSQR, on the HBOLBSV matrix in Table \ref{tb:matrixproperties}. The degree of the filter is $m = 11$ and the number of vectors is $k = 8$.}
}
\label{fig:component-scaling}
\end{figure}

\begin{figure}[ht!]
  \begin{center}
    \begin{tabular}{cc}
      \includegraphics[height=2.4in]{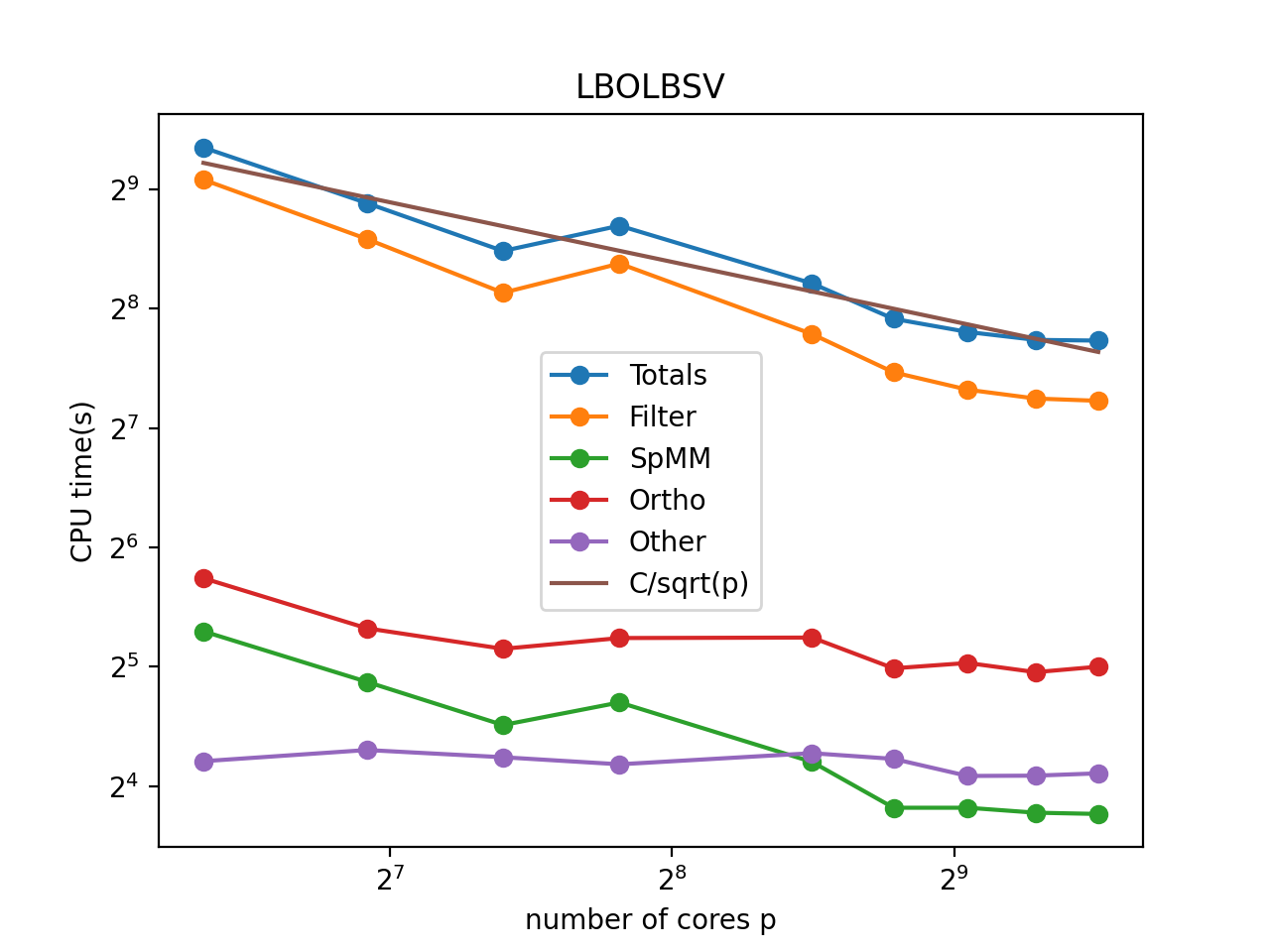} &
      \includegraphics[height=2.4in]{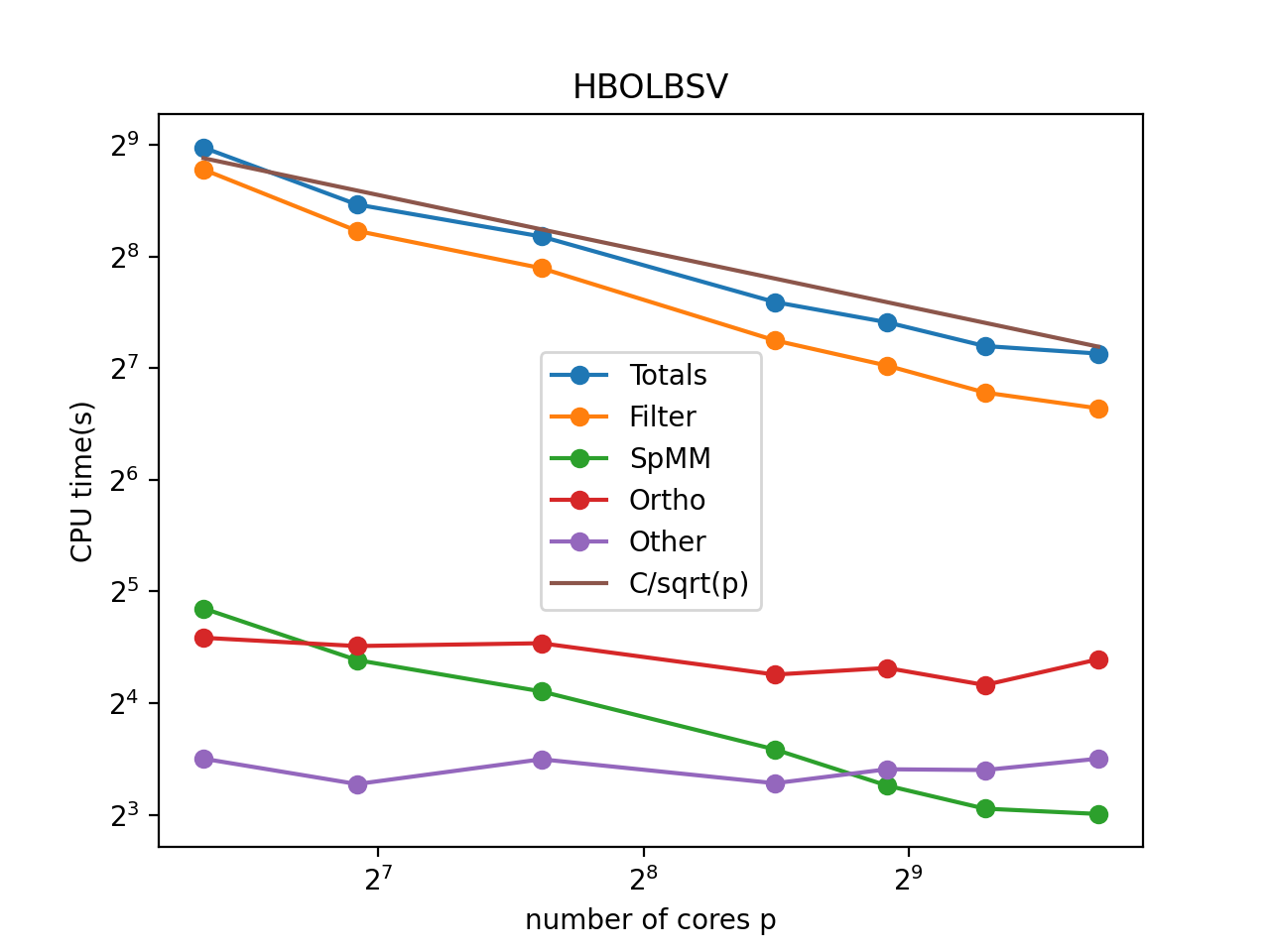} \\
      \includegraphics[height=2.4in]{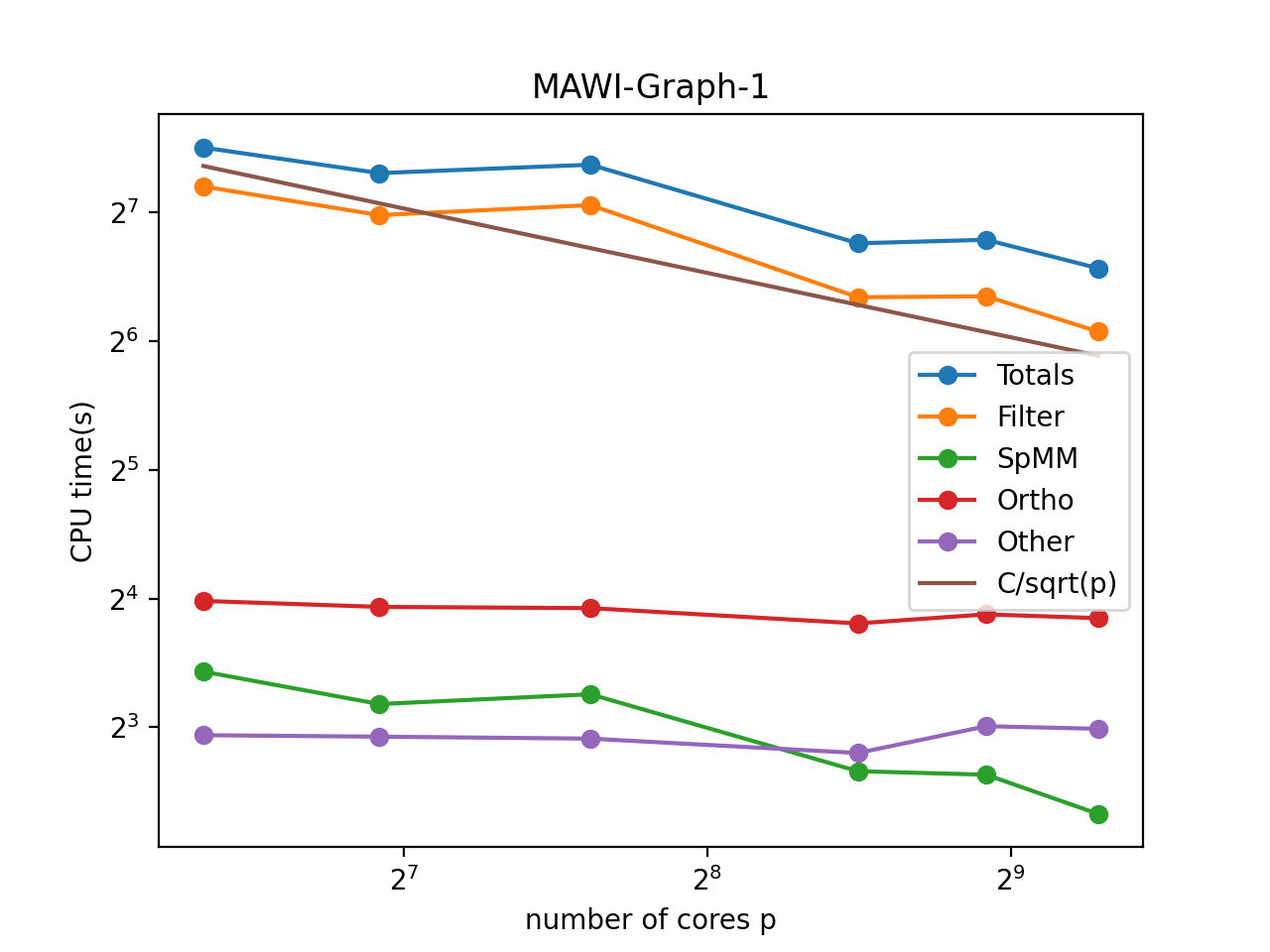} &
      \includegraphics[height=2.4in]{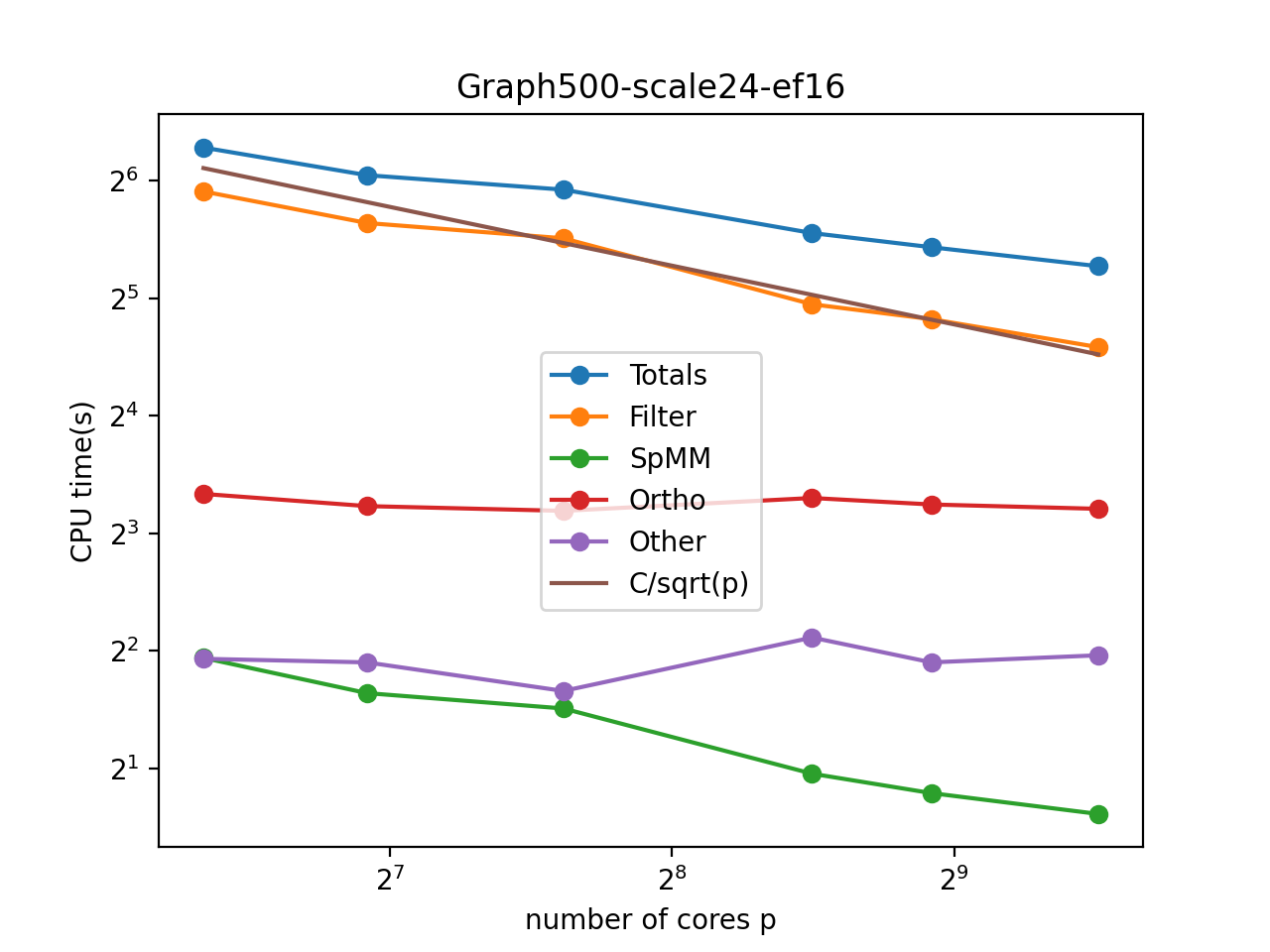}
    \end{tabular}
  \end{center}
\caption{\textbf{Scaling of the distributed Block Chebyshev-Davison algorithm and its components to numbers of cores. The stopping criteria are $tol = 10^{-3}$ and the degrees are $m = 15$. The number of eigenvectors $k$ and the number of vectors added per iteration $k_{b}$ vary for matrices. LBOLBSV: $k = 16, k_{b} = 16$; HBOHBSV: $k = 4, k_{b} = 4$; MAWI-Graph-1: $k = 4, k_{b} = 4$; Graph500-scale24-ef16: $k = 4, k_{b} = 4$.}
}
\label{fig:dbchdav-scaling}
\end{figure}

\begin{figure}[ht!]
  \begin{center}
    \begin{tabular}{c}
      \includegraphics[height=3.4in]{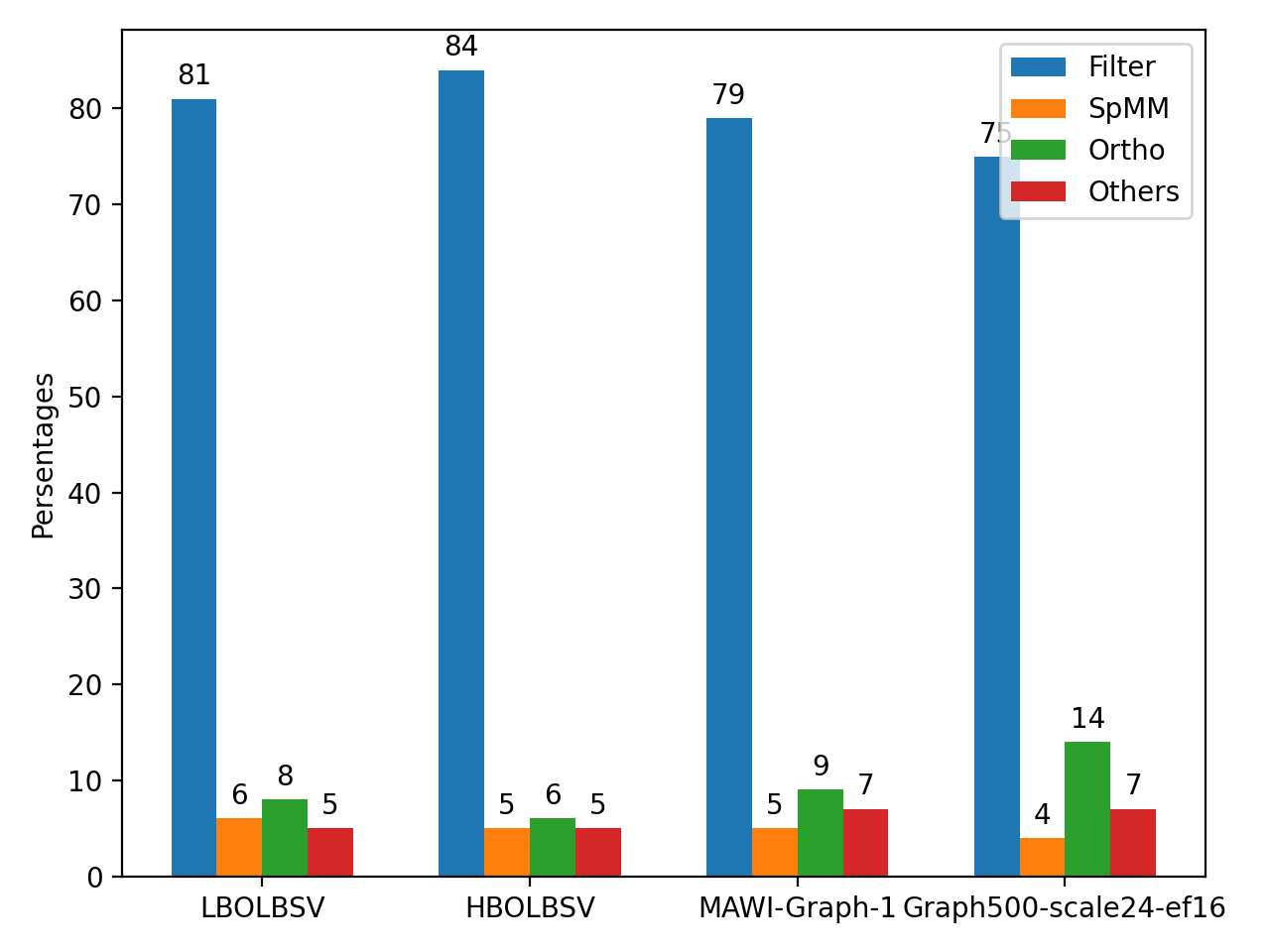} 
    \end{tabular}
  \end{center}
\caption{\textbf{Percentages of the CPU time spent on components in the experiments shown in Figure \ref{fig:dbchdav-scaling} when the number of cores $p = 121$. }
}
\label{fig:components-ratio}
\end{figure}

\begin{figure}[ht!]
  \begin{center}
    \begin{tabular}{ccc}
      \includegraphics[height=1.5in]{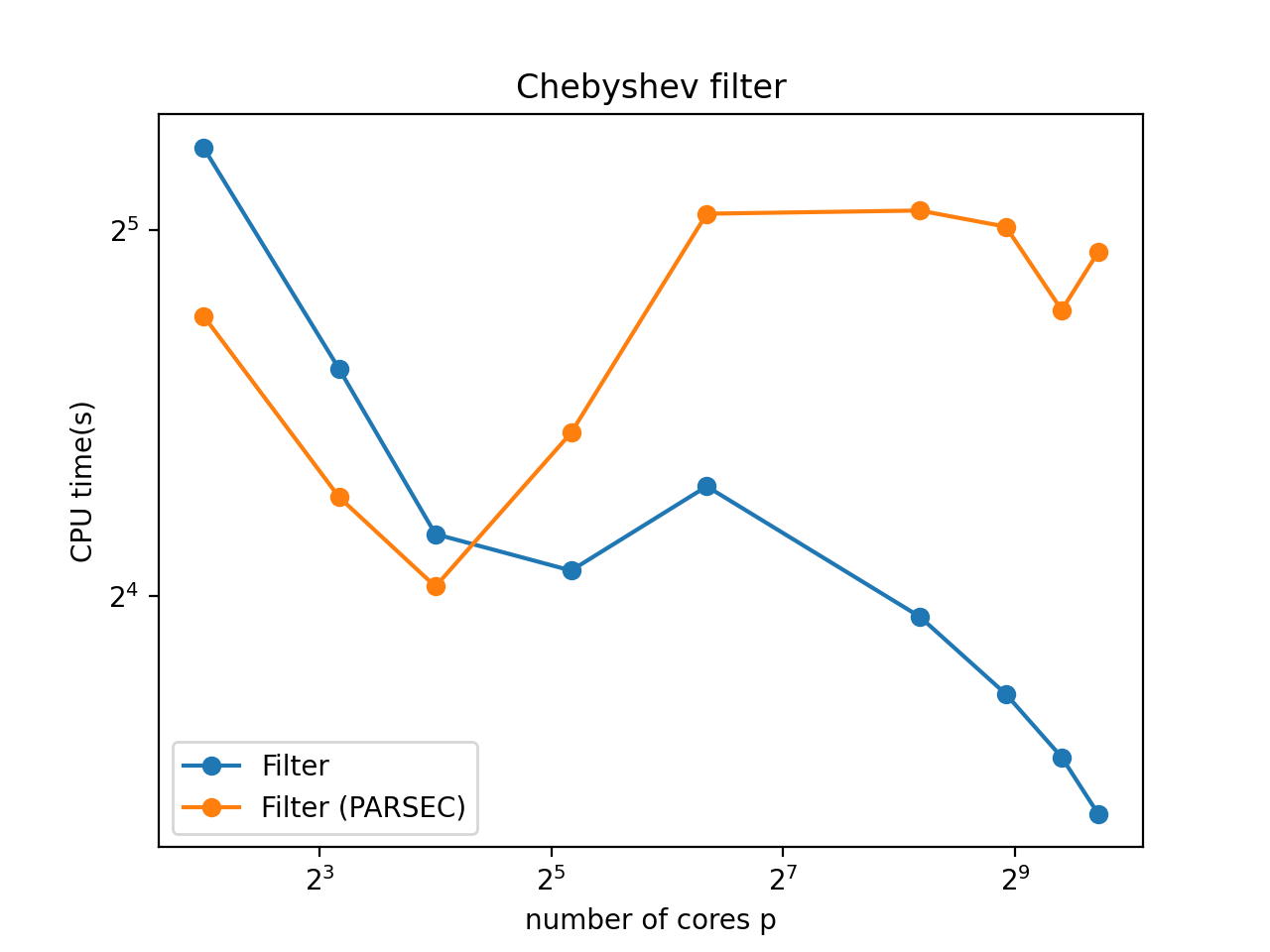} &
      \includegraphics[height=1.5in]{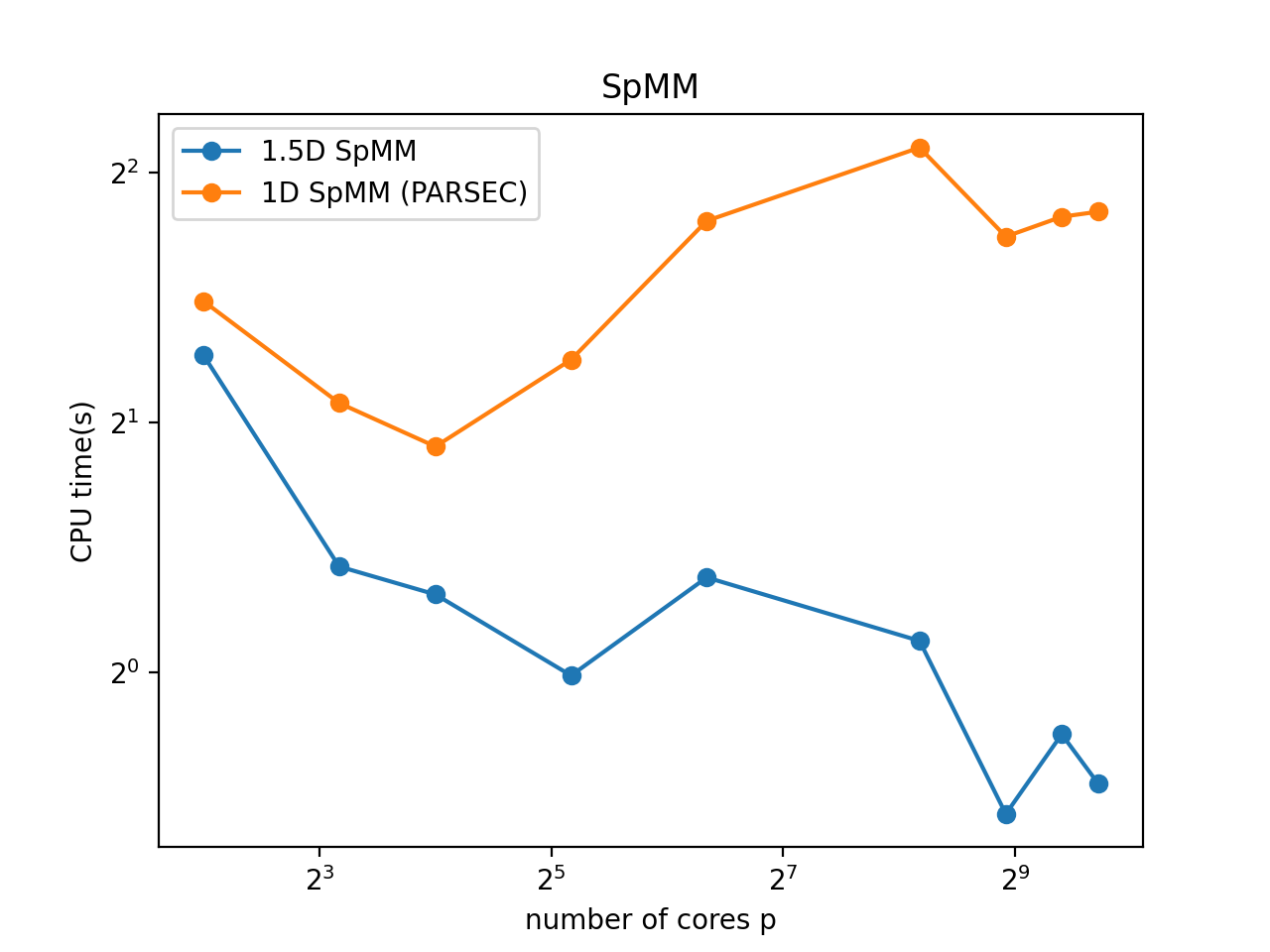} &
      \includegraphics[height=1.5in]{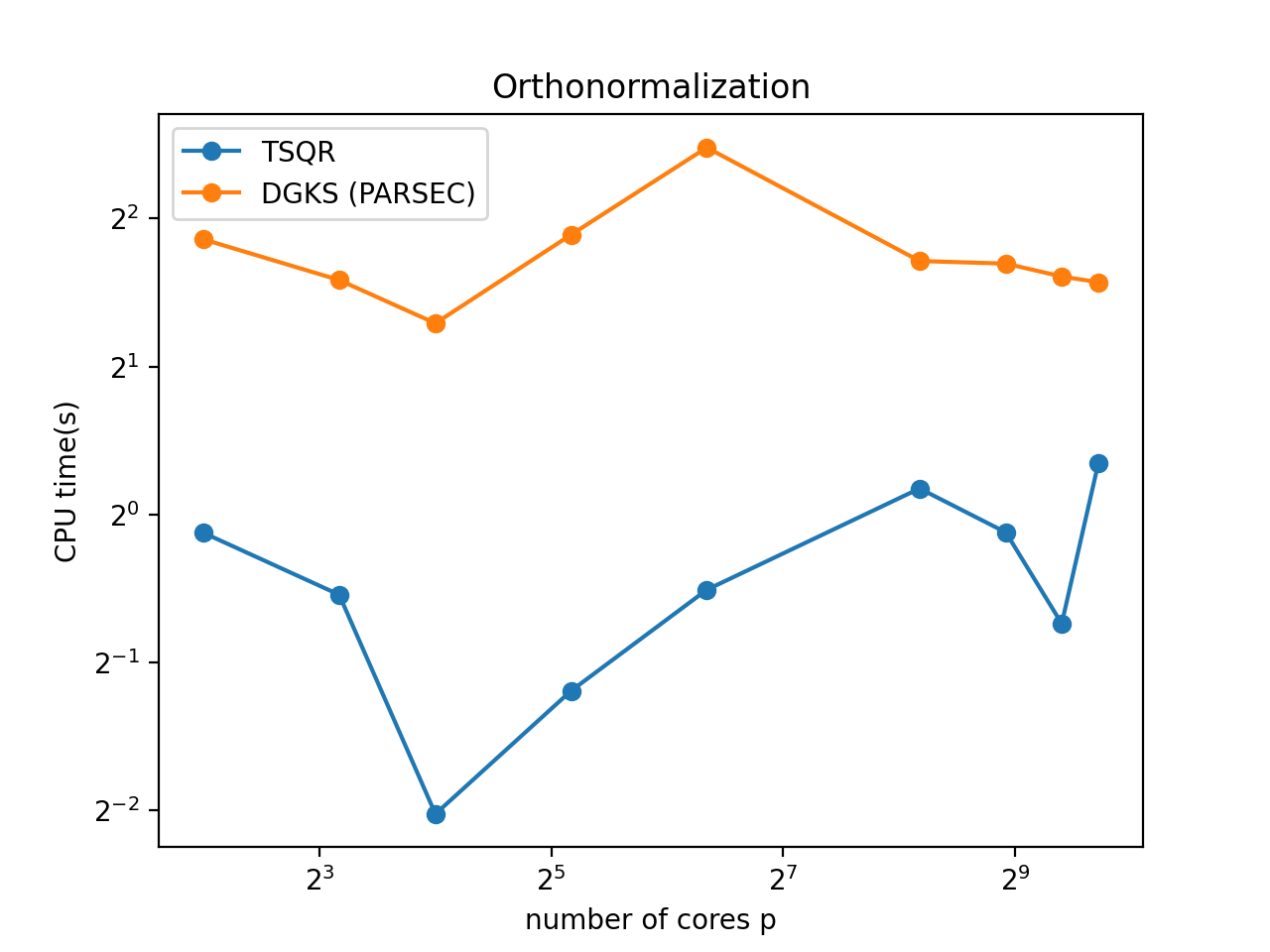} \\
    \end{tabular}
  \end{center}
\caption{\textbf{Comparison of our implementation and PARSEC's implementation of Chebyshev polynomial filters, SpMM, and orthonormalization. The experiment is conducted on the LBOLBSV matrix in Table \ref{tb:matrixproperties} with problem size $N = 5\times 10^6$, the number of vectors $k = 16$, and the degree of the filter $m=11$.}
}
\label{fig:dbchdav-comparison}
\end{figure}

\subsection{Scalability of parallel eigensolvers}
{In this experiment, we test the scalability of our parallel Block Chebyshev-Davidson method, ARAPCK, and LOBPCG. We first test the scalability of parallel ARPACK and LOBPCG via the PETSc library \cite{petsc-web-page, petsc-user-ref, petsc-efficient} on the graph LBOLBSV(SG)-1M. See Figure \ref{fig:scalability-arpack-lobpcg}. Parallel ARPACK and LOBPCG lose scalability when the number of processes exceeds 256. This is mainly because they carry out orthogonalization or orthonormalization at every iteration, which does not scale in parallel environments.

To show the scalability of our parallel Block Chebyshev-Davidson method, we consider four sparse normalized Laplacian matrices input, including traffic data from the MAWI Project (MAWI-Graph-1 \cite{cho2000traffic}), synthetic data at various scales generated using the scalable Graph500 Kronecker generator (Graph500-scale24-ef16 \cite{kepner2018design}), and two matrices from the Graph Challenge (LBOLBSV and HBOLBSV). 
Various properties of these matrices are presented in Table \ref{tb:matrixproperties}. Load imbalance is defined as the ratio of the maximum number of nonzeros assigned to a process to the average number of nonzeros in each process:
\begin{equation}
    \dfrac{p * \max_{i,j}nnz(A[i,j])}{nnz(A)}.
\end{equation}

We test the scaling of our distributed algorithm and its components on the four matrices. Figure \ref{fig:component-scaling} illustrates how local computation and communication in a Chebyshev polynomial filter, an SpMM, and an orthonormalization using TSQR scale to large concurrencies.
The speedup of the filter and SpMM is roughly proportional to $\sqrt{p}$ because communication is more costly in the two components and accelerated at a rate roughly proportional to $\sqrt{p}$. The orthonormalization using distributed TSQR does not scale well due to communication costs.   
Figure \ref{fig:dbchdav-scaling} presents the scaling of the distributed algorithm and its components and shows the speedup of the Chebyshev polynomial filter, and hence the entire algorithm is roughly proportional to $\sqrt{p}$. Though the distributed orthonormalization and other components, including updating the Rayleigh-quotient matrix and evaluating residuals, do not scale well to the numbers of cores, Figure \ref{fig:components-ratio} shows that they are minor, and the filters are dominant in terms of CPU time. Note that a higher degree of a Chebyshev polynomial filter tends to result in faster convergence \cite{zhou2007chebyshev, zhou2010block} and more dominance among other components. Therefore, the distributed algorithm is efficient and practically scalable to a large number of cores.}

\subsection{Comparison of different implementations}
We compare our distributed implementation of the Block Chebyshev-Davidson method with the distributed implementation in PARSEC. Comparing the corresponding implementations of SpMM, Chebyshev polynomial filters, and orthonormalization is sufficient. Figure \ref{fig:dbchdav-comparison} shows our implementations consistently outperform the implementations in PARSEC in efficiency and scalability. Indeed, the implementations in PARSEC do not scale to relatively large concurrencies. 

\section{Conclusion}
{ARPACK and LOBPCG are the most frequently used eigensolvers in spectral clustering. We propose to use the Block Chebyshev-Davidson method for eigenvector computation in spectral clustering. Due to the known analytic bounds of the spectrum of the symmetric normalized Laplacian of an undirected graph, one does not need prior estimation of the bounds in the vanilla Block Chebyshev-Davidson method. Furthermore, the exact analytic bounds also help the convergence of the method. Numerical results show that the Block Chebyshev-Davidson method produces competitive clustering quality. Though it is a bit slower than ARPACK and LOBPCG, it is more scalable in parallel computing environments, which means it has more potential to handle spectral clustering for large graphs. Although all three eigensolvers conduct orthogonalization at each iteration, orthogonalization only takes a small portion in the sense of computation costs in the parallel Block Chebyshev-Davidson method due to the use of the Chebyshev polynomial filter. Since a higher degree of a Chebyshev polynomial filter results in faster convergence and more dominance among other components, the parallel algorithm is efficient and practically scalable to a large number of cores.}

% We propose a distributed Block Chebyshev-Davidson method for spectral analysis in spectral clustering. One of the advantages of applying the Block Chebyshev-Davidson method in such a spectral analysis is that the theoretically optimal lower and upper bounds are known, so the method does not need to estimate the bounds and converges fast. Numerical results show that the sequential method is at least comparable to LOBPCG. The distributed Block Chebyshev-Davidson algorithm is developed based on an A-Stationary 1.5D SpMM algorithm and a parallel TSQR. Though the distributed orthonormalization using TSQR does not scale well, the distributed Chebyshev polynomial filters and SpMMs scale well and dominate the entire algorithm in terms of running time. The SpMM, filters, and hence the entire algorithm are practically scalable to a large number of processes and accelerated approximately at a rate $\sqrt{p}$. Our new distributed algorithm consistently outperforms its counterpart in PARSEC in both efficiency and scalability.

\section{Acknowledgments}
We thank Aleksey Urmanov for helpful discussion and comments, and thank Oracle Labs, Oracle Corporation, Austin, TX, for providing funding that supported research in the area of scalable spectral clustering and distributed eigensolvers. H. Y. was partially supported by the US National Science Foundation under awards DMS-2244988, DMS-2206333, and the Office of Naval Research Award N00014-23-1-2007. 

\bibliographystyle{unsrt} 
\bibliography{ref}
\end{document}